\documentclass{article}

\usepackage{PRIMEarxiv}

\usepackage[utf8]{inputenc} 
\usepackage[T1]{fontenc}    
\usepackage{hyperref}       
\usepackage{url}            
\usepackage{booktabs}       
\usepackage{amsfonts}       
\usepackage{nicefrac}       
\usepackage{microtype}      
\usepackage{lipsum}
\usepackage{fancyhdr}       
\usepackage{graphicx}       
\usepackage{adjustbox}
 \usepackage{enumitem,color}
\usepackage{amsmath,amssymb,bbm,amsthm}
\usepackage{subcaption}
\usepackage{tabularx}
\usepackage{xcolor,array}         
\usepackage{multirow,colortbl}
\usepackage[linesnumbered,ruled,vlined]{algorithm2e}
\usepackage{algpseudocode}

\usepackage[frozencache=true,cachedir=minted-cache]{minted} 

\SetKwFunction{Push}{Push}
\SetKwFunction{Append}{Append}
\SetKwFunction{Yield}{Yield}
\SetKwFunction{Pop}{Pop}
\SetKwFunction{Return}{Return}
\SetKwFunction{Sort}{Sort}
\SetKwFunction{Add}{Add}
\SetKwFunction{Remove}{Remove}

\newcolumntype{Y}{>{\centering\arraybackslash}X}

\title{Generative artificial intelligence for \\
navigating synthesizable chemical space
}

\author{
  Wenhao Gao$^{*}$ \\
  Department of Chemical Engineering \\
  Massachusetts Institute of Technology \\
  Cambridge, MA 02139, USA\\
  \texttt{whgao@mit.edu} \\
   \And
  Shitong Luo$^{*}$ \\
  Department of Electrical Engineering and Computer Science \\
  Massachusetts Institute of Technology \\
  Cambridge, MA 02139, USA\\
  \texttt{luost@mit.edu} \\
  \And
  Connor W. Coley \\
  Department of Chemical Engineering \\
  Department of Electrical Engineering and Computer Science \\
  Massachusetts Institute of Technology \\
  Cambridge, MA 02139, USA\\
  \texttt{ccoley@mit.edu} \\
  \\
  $^*$Equal Contribution \\
}

\begin{document}
\maketitle

\begin{abstract}
We introduce SynFormer, a generative modeling framework designed to efficiently explore and navigate synthesizable chemical space. Unlike traditional molecular generation approaches, we generate synthetic pathways for molecules to ensure that designs are synthetically tractable. By incorporating a scalable transformer architecture and a diffusion module for building block selection, SynFormer surpasses existing models in synthesizable molecular design. We demonstrate SynFormer's effectiveness in two key applications: (1) local chemical space exploration, where the model generates synthesizable analogs of a reference molecule, and (2) global chemical space exploration, where the model aims to identify optimal molecules according to a black-box property prediction oracle. Additionally, we demonstrate the scalability of our approach via the improvement in performance as more computational resources become available. With our code and trained models openly available, we hope that SynFormer will find use across applications in drug discovery and materials science.
\end{abstract}

\section{Introduction}
\label{sec:intro}

The discovery of novel functional molecules is a central challenge in chemical science and engineering and is crucial for addressing key societal and technological challenges, including those related to healthcare \cite{walters2011medicinal,wu2016small,lyu2019ultra}, energy \cite{yu2020molecular,hachmann2011harvard}, and sustainability \cite{rai2021recent,diederichsen2022electrochemical,peng2022human}. However, the process of discovery is often risky, complex, time-consuming, and resource-intensive \cite{pyzer2015high,smietana2016trends,pushpakom2019drug}. Recent advances in artificial intelligence (AI) \cite{lecun2015deep}, particularly in generative modeling \cite{wang2019variational,vaswani2017attention,ho2020denoising}, have opened up new avenues for generative molecular design \cite{sanchez2018inverse,gomez2018automatic,jin2018junction,blaschke2020reinvent,fu2021differentiable,meyers2021novo,anstine2023generative,subramanian2024closing}. The expressivity of deep learning models, coupled with automatic differentiation \cite{paszke2017automatic} and increasingly affordable computational resources, have made it possible to model the complex distribution of molecular structures directly. This capability allows for efficient exploration of broader virtual chemical space than traditional virtual screening approaches \cite{walters1998virtual,shoichet2004virtual,walters2018virtual}, with more fine-grained and controllable navigation compared to conventional combinatorial fragment-based methods \cite{venkatasubramanian1995evolutionary,jensen2019graph}. 

Despite the promise of generative design methods, their adoption has remained somewhat limited. One major challenge is that most generative models often produce synthetically intractable molecular structures when targeting specific design goals, such as optimizing property scores \cite{gao2020synthesizability,renz2019failure,walters2021critical,stanley2023fake}. When designed molecules cannot be synthesized and validated in the lab at a reasonable cost, their practical value is limited. In addition, the actionability of computer-aided molecular design is crucial for achieving rapid design cycles in early-stage drug discovery \cite{nicolaou2019idea2data,brocklehurst2024microcycle} and for enabling closed-loop autonomous discovery \cite{schneider2018automating,coley2020autonomous1,coley2020autonomous2,koscher2023autonomous,strieth2024delocalized}.
This issue, though long-standing and observed in earlier combinatorial algorithms, has become more pronounced with the advent of deep generative AI models.  

There has been considerable interest in incorporating synthetic accessibility into molecular design methods \cite{gao2020synthesizability,stanley2023fake}. The most straightforward strategy is to incorporate a heuristic synthetic accessibility score as a design criterion \cite{gao2020synthesizability,seumer2023computational}. However, quantitatively estimating synthetic accessibility must account for factors such as regioselectivity, functional group compatibility, and building block availability, all of which contribute to a rugged structure-synthesizability landscape that makes the design of such scores an ongoing challenge. While notable efforts have been made to quantify synthetic accessibility \cite{ertl2009estimation,thakkar2021retrosynthetic,liu2022retrognn}, achieving reliable quantification through a simple heuristic remains a distant goal. Advances in the sample efficiency of molecular optimization \cite{gao2022sample,guo2024saturn} have made it feasible to conduct explicit retrosynthesis analysis for each designed molecule as a means of evaluating synthetic accessibility \cite{guo2024directly}. However, the computational overhead remains significant, making this approach impractical for most existing generative design models. Furthermore, unsynthesizable molecules provide no learning signal, and when a model continues to propose such molecules, the sparse feedback hinders effective learning. 

A more ideal and effective approach to synthesizable molecular design, in our opinion, involves constraining the design process to focus exclusively on synthesizable molecules by designing \emph{synthetic pathways} rather than simply designing \emph{structures}. With the growing role of massive make-on-demand libraries—whose sizes exceed what can be fully enumerated \cite{walters2018virtual,hoffmann2019next,patel2020savi,grygorenko2020generating,WuXi_AppTec_2022,neumann2023relevance,bedart2024pan}-this class of methods has gained increasing interest, spanning both traditional Monte Carlo techniques \cite{vinkers2003synopsis,hartenfeller2012dogs,korovina2020chembo,nguyen2021generative,sadybekov2022synthon,swanson2024generative,sun2024syntax} and more recent generative approaches \cite{bradshaw2019model,gottipati2020learning,horwood2020molecular,bradshaw2020barking,gao2021amortized,koziarski2024rgfn,luo2024projecting}. Despite this progress, current synthesis-centric methods still fall short in terms of controllability and efficiency in navigating the chemical space, and as a result have not achieved widespread adoption. 
These methods that rely on property scores to guide the design process often exhibit lower efficiency in optimizing expensive oracles compared to structure-centric methods \cite{gao2022sample}, as they model the synthetic action sequence-property landscape, which is inherently more complex than the structure-property landscape. Some recent models offer a more controlled exploration of local chemical spaces by using a reference input molecule \cite{bradshaw2019model,bradshaw2020barking,gao2021amortized,luo2024projecting}. However, the low reconstruction rates for theoretically feasible molecules suggest a risk of partial collapse during decoding, suggesting that the practical chemical space accessible to these models is smaller than the theoretically accessible space. This behavior leads to certain regions of the chemical space becoming inaccessible, regardless of the input, ultimately hindering the overall design process.

Herein, we build upon our previous contributions \cite{gao2021amortized,luo2024projecting} to introduce SynFormer, a generative AI framework designed for efficient and controllable navigation within a synthesizable chemical space. Like earlier synthesizable design methods, SynFormer is a synthesis-centric approach, generating synthetic pathways by readily available building blocks through robust chemical transformations, ensuring synthetic traceability subject to the limitations of those transformation rules. SynFormer distinguishes itself from previous models with its compute-efficient and scalable transformer architecture \cite{vaswani2017attention}, which empirically shows improvements in performance as the training dataset grows. To select suitable molecular building blocks from the large, discrete, and multimodal space of commercially-available options, SynFormer incorporates a denoising diffusion model as a token head module \cite{ho2020denoising}. The framework is fully end-to-end differentiable, enabling effective training and optimization.

We demonstrate the utility and versatility of the SynFormer framework through two instantiations: (1) SynFormer-ED, an encoder-decoder model that generates synthetic pathways corresponding to a given input molecule for exact or approximate reconstruction of that input, and (2) SynFormer-D, a decoder-only model for generating synthetic pathways that is amenable to fine-tuning towards specific property goals. Both models are trained on a simulated chemical space derived from a curated set of 115 reaction templates and 223,244 commercially available building blocks, extending beyond Enamine’s REAL Space \cite{grygorenko2020generating}. We validate SynFormer’s capabilities by demonstrating success on (a) reconstructing molecules within both the Enamine REAL and ChEMBL \cite{mendez2019chembl} chemical spaces; (b) local synthesizable chemical space exploration given a reference molecule; and (c) global synthesizable chemical space exploration guided by black-box property prediction model. These results not only highlight SynFormer's ability to navigate synthesizable chemical space using varied control strategies but also demonstrate its practical applicability in real-world molecular design use cases. The scalability of the SynFormer framework with respect to both training data and model size suggests considerable potential for further performance enhancement. In conclusion, these findings underscore SynFormer’s flexibility and potential to impact molecular design across various domains, including drug development and materials science.

\section{SynFormer: a generative framework for synthesizable molecular design}
\label{sec:synformer}

\begin{figure*}[th!]
\centering
\includegraphics[width=0.85\linewidth]{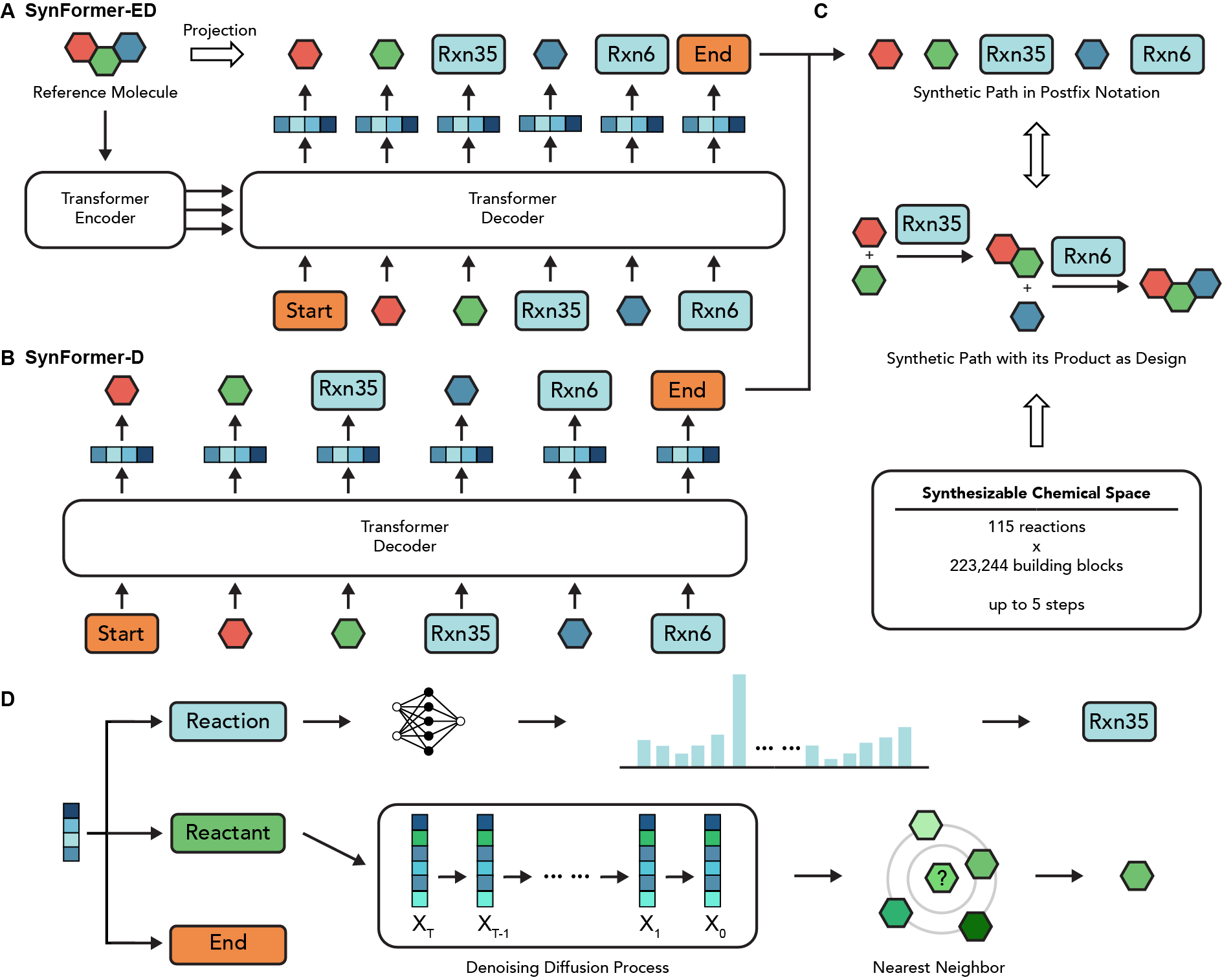}
\caption{Schematic illustration of the SynFormer framework and architecture. (A) The SynFormer-ED architecture is an encoder-decoder that takes a molecule as input and outputs a synthetic route to the same or an analogous molecule. (B) SynFormer-D is a decoder-only framework designed to generate synthetic routes. (C) Synthetic routes are tokenized using a postfix notation to make them amenable to autoregressive generation. Routes are constructed by applying 115 reactions to a set of 223,244 molecular building blocks, covering a synthesizable chemical space estimated as $>10^{60}$ molecules. (D) During generation, a token generated by the transformer is first classified by token type. If the token represents a reaction or reactant (building block), it undergoes an additional classification process to select the appropriate reactions or a denoising diffusion process followed by a nearest neighbor search to select the appropriate building block(s).}
\label{fig:schematics}
\end{figure*}

SynFormer is a generative framework designed for modeling synthesizable chemical spaces, as schematically illustrated in Figure \ref{fig:schematics}. Although it is challenging to predict which molecules can be synthesized without experimental validation, the progress of make-on-demand libraries such as Enamine REAL Space \cite{grygorenko2020generating}, GalaXi \cite{WuXi_AppTec_2022}, and eXplore \cite{neumann2023relevance} has demonstrated that molecules constructed by connecting purchasable molecular building blocks through a series of curated, reliable reactions have a high likelihood of being synthesizable. Therefore, within the scope of this work, we define synthesizable chemical space as encompassing all molecules that can be formed within the reaction network by linking purchasable building blocks through up to five steps of known chemical transformations, though in principle, there is no limit to the length of pathways that can be generated, and building blocks and reactions beyond our selected list can also be utilized.

To effectively model chemically relevant and practically useful spaces, we adapted the reaction set used to construct REAL Space, which primarily focuses on bi- and tri-molecular reactions that couple multiple building blocks together. We further augment this set with additional common reactions, such as functional group interconversions, to better simulate a general organic synthesis endeavor, resulting in a total of 115 reaction templates. As the set of purchasable molecular building blocks, we use Enamine's U.S. stock catalog to ensure high availability, which also approximates those used to generate REAL Space. Although the proportion of synthesizable molecules may vary due to the broader focus of the chemical space, our model theoretically covers a chemical space broader than then tens of billions currently represented in the Enamine REAL Space, making it suitable for extensive exploration during the molecular design stage. Both the choice of templates and the choice of building blocks can be straightforwardly modified prior to retraining and are not inherent to SynFormer's approach.

Generating synthetic pathways requires careful consideration of their data structure. Bottom-up synthesis planning involves sequential decision-making with branches and merges, encompassing data modalities that span reaction templates and molecular building blocks, which represent small and very large discrete design spaces, respectively. To manage this complexity, we adopt the postfix notation to represent synthetic pathways linearly, as described in  Luo et al.'s \cite{luo2024projecting} (Figure \ref{fig:schematics}C). Specifically, we use four types of tokens to represent synthetic paths: a start token [START], an end token [END], reaction tokens [RXN], and building block tokens [BB]. Similar to the postfix notation of mathematical formulae, the reactions are placed after the reagents, allowing the sequence to be processed in a step-by-step manner without loss of generality in the synthetic pathways (e.g., the ability to accommodate any linear or convergent sequence). This linear notation also enables the use of autoregressive decoding via a transformer architecture \cite{vaswani2017attention}, which is widely recognized as a scalable model backbone. A stack of standard transformer layers processes the sequence as it is decoded, taking the previous tokens and outputting the embedding for the next token. Each token embedding then passes through a multi-layer perceptron (MLP) classification head to determine the token type. If the token is [END], the generation process terminates. If the token is [RXN] or [BB], the embedding is used to select which building blocks to add and which reaction type to perform.

The selection of reactions can be relatively straightforward using an additional classification head; however, the number of purchasable building blocks can easily exceed hundreds of thousands, millions, or perhaps billions, and will continue to grow over time. Instead of relying on a static classification network, our past work \cite{gao2021amortized,luo2024projecting} has involved generating Morgan fingerprints \cite{morgan1965generation} and then retrieving the nearest building block from a list of candidates to enable generalization to unseen building blocks. To further enhance this building block selection process, we adopt a denoising diffusion probabilistic module \cite{ho2020denoising} to predict the posterior distribution of molecular fingerprints conditioned on the token embedding. Specifically, we model the molecular fingerprints using an $n$-dimensional joint Bernoulli distribution (where $n$ is the length of the fingerprint) and employ a multinomial diffusion framework \cite{hoogeboom2021argmax} to handle the discrete nature of the data. Using the denoising diffusion module as the building block token head offers several advantages: while maintaining the end-to-end differentiable nature, it allows for scaling to longer fingerprints (2048 in this work compared to 256 in previous studies), thereby avoiding bit collisions that can occur with shorter fingerprints and enabling generation at a finer resolution. Additionally, this approach more effectively handles multimodal distributions, where multiple ``true" answers exist for a single decision--a particularly common scenario when modeling synthetic pathways.

Based on this generative modeling framework, we implemented two specific models: SynFormer-D, which straightforwardly implements a transformer decoder with the aforementioned reaction classification heads and building block diffusion heads (Figure \ref{fig:schematics}B), and SynFormer-ED, which adds a standard transformer encoder to condition generation on an input SMILES string \cite{weininger1988smiles} (Figure \ref{fig:schematics}A). To train the models, we uniformly sample synthetic pathways from the synthesizable chemical space simulated from the reaction templates and building block list as training data. Detailed descriptions of the model architectures, training protocols, and hyperparameter settings are provided in the Methods section and Supplementary Information.

\section{Results}
\label{sec:results}

\subsection{Molecule reconstruction and chemical space coverage}
\label{sec:reconstruction}

An important factor that affects a model's performance in molecular design applications is its coverage of chemical space \cite{zhang2021comparative}. Even though the theoretical synthesizable chemical space constructed from reaction templates and building blocks is vast, previous models have shown that a trained model can struggle to generate molecules that are known to be synthesizable \cite{bradshaw2020barking,gao2021amortized,nguyen2021generative,luo2024projecting}. This limitation can hinder the model's effectiveness in molecular design if it cannot access the chemical space where the optimal design resides. To assess the coverage of chemical space, we tested the SynFormer-ED model on molecule reconstruction: how often SynFormer-ED can reconstruct a viable synthetic path given a large set of potential molecules of interest.

\begin{figure}[th!]
\centering
\includegraphics[width=0.85\linewidth]{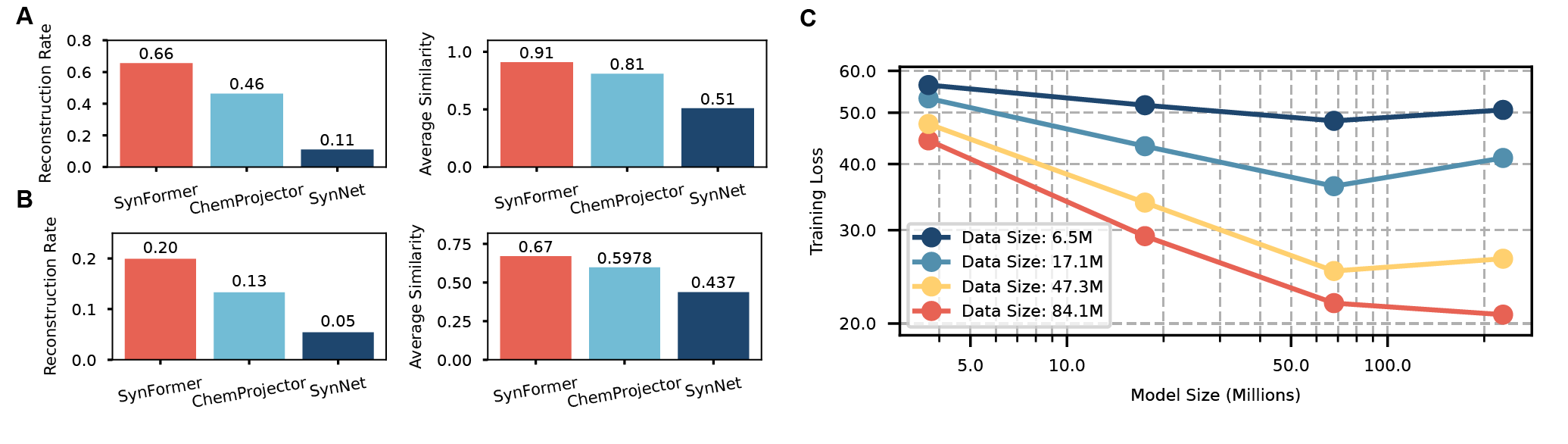}
\caption{Model performance on molecular reconstruction. (A and B) Comparison of the reconstruction rate and average structural (Tanimoto) similarity between input and output molecules for SynFormer-ED, ChemProjector \cite{luo2024projecting}, and SynNet \cite{gao2021amortized} on 1,000 randomly selected molecules from (A) REAL Diversity Set and (B) ChEMBL Database. (C) 
Scaling of model performance is measured by the training loss (binary cross-entropy (BCE) of the molecular fingerprint (FP) prediction) as model size and training data size increase.}
\label{fig:reconstruction}
\end{figure}

We evaluated reconstruction on a randomly selected subset of 1,000 molecules from ChEMBL Database and Enamine's REAL Diversity Set (Figure \ref{fig:reconstruction}A and B). SynFormer-ED successfully recovers 66\% of molecules from REAL, notably higher than that achieved by ChemProjector \cite{luo2024projecting} and SynNet \cite{gao2020synthesizability}. ChEMBL, by comparison, is a curated database of bioactive molecules with drug-like properties, which includes molecules whose syntheses may require building blocks or reactions beyond SynFormer's training set. This results in a reconstruction rate of 20\%---lower than in REAL Space but still an improvement over previous models. 

We further demonstrate that the model's performance, as measured by the binary cross-entropy (BCE) of the molecular fingerprints (FPs) \cite{morgan1965generation}, improves with increasing model size and training data (Figure \ref{fig:reconstruction}C). Notably, we also observed that the performance of the largest model did not outperform the second-largest model when insufficient data was available. This finding suggests that estimating training performance based on early epochs may not be an effective strategy when the model size surpasses a certain threshold \cite{frey2023neural}. These results emphasize the scalability of the architecture \cite{kaplan2020scaling} and indicate that further performance improvements can be achieved by increasing both the amount of data and computational resources. They also underscore the importance of scaling model size and training data in tandem to fully realize performance gains. As SynFormer is trained using simulated pathways according to its building blocks and reaction rules, there is virtually no limit to the size of training data available.

\subsection{Local chemical space exploration with SynFormer}
\label{sec:local}

In this section, we present the application of SynFormer-ED in exploring the local chemical space around a given reference molecule, projecting it into a synthesizable chemical space. We demonstrate two use cases of our model in molecular design: (1) generating synthesizable analogs for unsynthesizable designs and (2) hit expansion within synthesizable chemical space. 

\begin{figure*}[th!]
\centering
\includegraphics[width=0.85\linewidth]{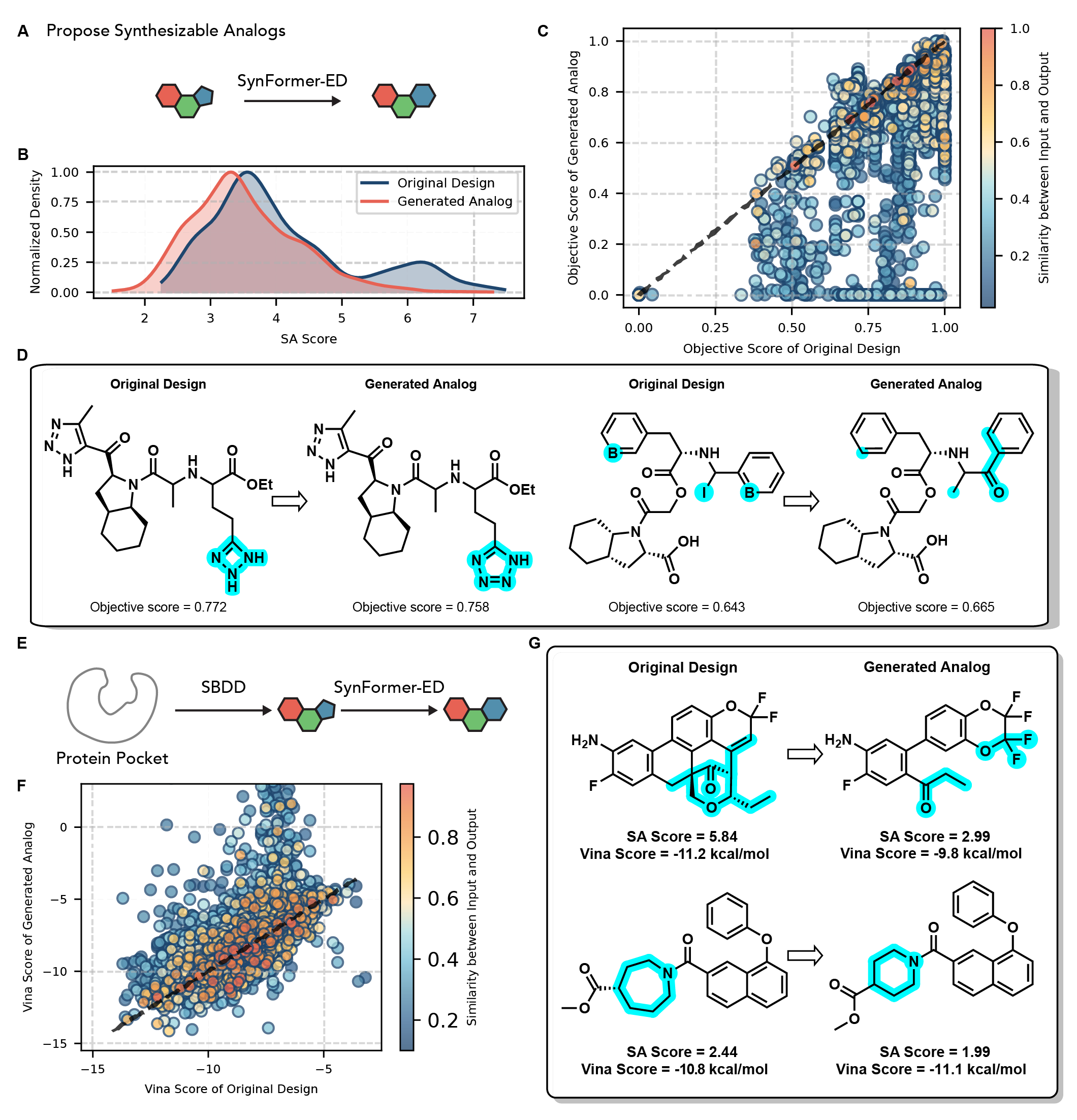}
\caption{Application of SynFormer in projecting unsynthesizable design into synthesizable chemical space. (A) Schematic illustration of SynFormer-ED generating synthesizable structural analogs for unsynthesizable molecules. (B) Normalized distribution of SA scores for the originally designed molecules and their corresponding synthesizable analogs. Note that the distributions are normalized to a peak value of 1 for a clearer comparison. (C) Scatter plot comparing objective scores of originally designed molecules versus their generated analogs. Points are colored by the structural similarity between them, showing that structurally similar analogs tend to possess close properties. (D) Examples of originally designed unsynthesizable molecules and their SynFormer-generated synthesizable analogs. Objective scores are shown beneath each molecule, demonstrating comparable activities for the generated analogs. The modified fragments or atoms are highlighted in light blue. (E) The workflow shows SynFormer-ED generating synthesizable analogs for ligands generated by structure-based drug design. (F) Scatter plot comparing the Vina docking scores of originally designed ligands and their generated analogs. Points are colored based on the structural (Tanimoto) similarity between the input and output, showing strong agreement in general and an ability to generate analogs with comparable scores. (G) The original design and generated analog for Estrogen receptor alpha, with their SA Score and Vina score below. The modified fragments or atoms are highlighted in light blue.}\label{fig:projection}
\end{figure*}

\subsubsection{Synthesizable analog generation}
\label{sec:analog}

In molecule reconstruction experiments, we observed that SynFormer-ED is able to generate structurally similar outputs even when provided with unsynthesizable (according to its building blocks and reactions) molecules. This capability allows SynFormer-ED to be applied to generate synthesizable structural analogs of unsynthesizable designs, as illustrated in Figure \ref{fig:projection}A. The implicit goal is to preserve as much of the overall structure and key pharmacophores as possible, effectively transforming non-feasible molecules into synthesizable compounds, akin to ``projecting'' the input molecule onto the synthesizable space.

To evaluate the impact of synthesizable projection, we first followed the experimental setup of Luo et al. \cite{luo2024projecting} and used SynFormer-ED to generate synthesizable analogs for molecules identified as unsynthesizable by ASKCOS \cite{askcos} in \cite{gao2020synthesizability}. The original molecules were optimized using \textit{de novo} design methods targeting ten multi-objective scores, each ranging from 0 to 1, where higher values are preferable. These scores encompass structural features, physicochemical properties, and metrics related to similarity or dissimilarity to known drugs, as well as the presence of specific substructures \cite{brown2019guacamol}. After the generation of analogs, we evaluated them using the same objective scores that the original designs were optimized for. As shown in Figure \ref{fig:projection}B, the synthesizable projection effectively eliminated the peak in synthetic accessibility (SA) scores \cite{ertl2009estimation} at around 6-7 observed in the original designs, shifting the distribution toward more easily synthesizable molecules. Additionally, a significant fraction of the generated analogs exhibited objective scores comparable to those of the original molecules (Figure \ref{fig:projection}C and Figure S2 in the Supporting Information). We demonstrate two examples in Figure \ref{fig:projection}D. Starting with high-scoring but seemingly-unsynthesizable molecules, SynFormer-ED generated analogs that preserved key structural features while correcting the unsynthesizable fragments. The generated analogs retained objective scores comparable to those of the original molecules and are inherently synthesizable (Figure S4 in Supporting Information), demonstrating SynFormer-ED’s ability to transform an unsynthesizable molecule into a synthesizable analog while maintaining structural integrity.

To further validate the utility of SynFormer-ED in realistic drug discovery scenarios, we explored its application in structure-based drug design (SBDD) \cite{gillet1994sprout,bohacek1996art,wang2000ligbuilder}. While numerous SBDD algorithms can generate ligands with promising predicted binding affinities (e.g., via 3D pocket-conditioned generation), these methods are often criticized for overlooking synthetic accessibility \cite{gillet1995sprout,harris2023benchmarking}. SynFormer-ED addresses this gap by projecting the outputs of such models into synthesizable chemical space (\ref{fig:projection}E). We demonstrate this approach using Pocket2Mol \cite{luo20213d} as an exemplary SBDD generative model.

Targeting 15 protein targets from the LIT-PCBA dataset \cite{tran2020lit}, we used SynFormer-ED to generate synthesizable analogs of molecules initially designed by Pocket2Mol. SynFormer-ED successfully produced analogs that preserve  similar Vina docking scores \cite{trott2010autodock} while ensuring that each design has a synthetic pathway (Figure \ref{fig:projection}F). Two examples are shown in Figure \ref{fig:projection}G: in the first example, the original design had a Vina score of -11.2 kcal/mol but a high synthetic accessibility (SA) score of 5.84, indicating poor synthesizability. SynFormer-ED generated a structurally similar analog with an improved SA score of 2.99 while exhibiting a Vina score of -9.8 kcal/mol. In the second example, although the original design was likely synthesizable with access to the appropriate building blocks, SynFormer-ED further reduced the SA score from 2.44 to 1.99 with minor improvements in Vina score. These results demonstrate SynFormer-ED’s potential to complement existing SBDD algorithms by generating more practical and synthesizable drug candidates, suitable for experimental validation and further development.

\subsubsection{Hit expansion within synthesizable chemical space}
\label{sec:hit}

\begin{figure*}[ht!]
\centering
\includegraphics[width=0.85\linewidth]{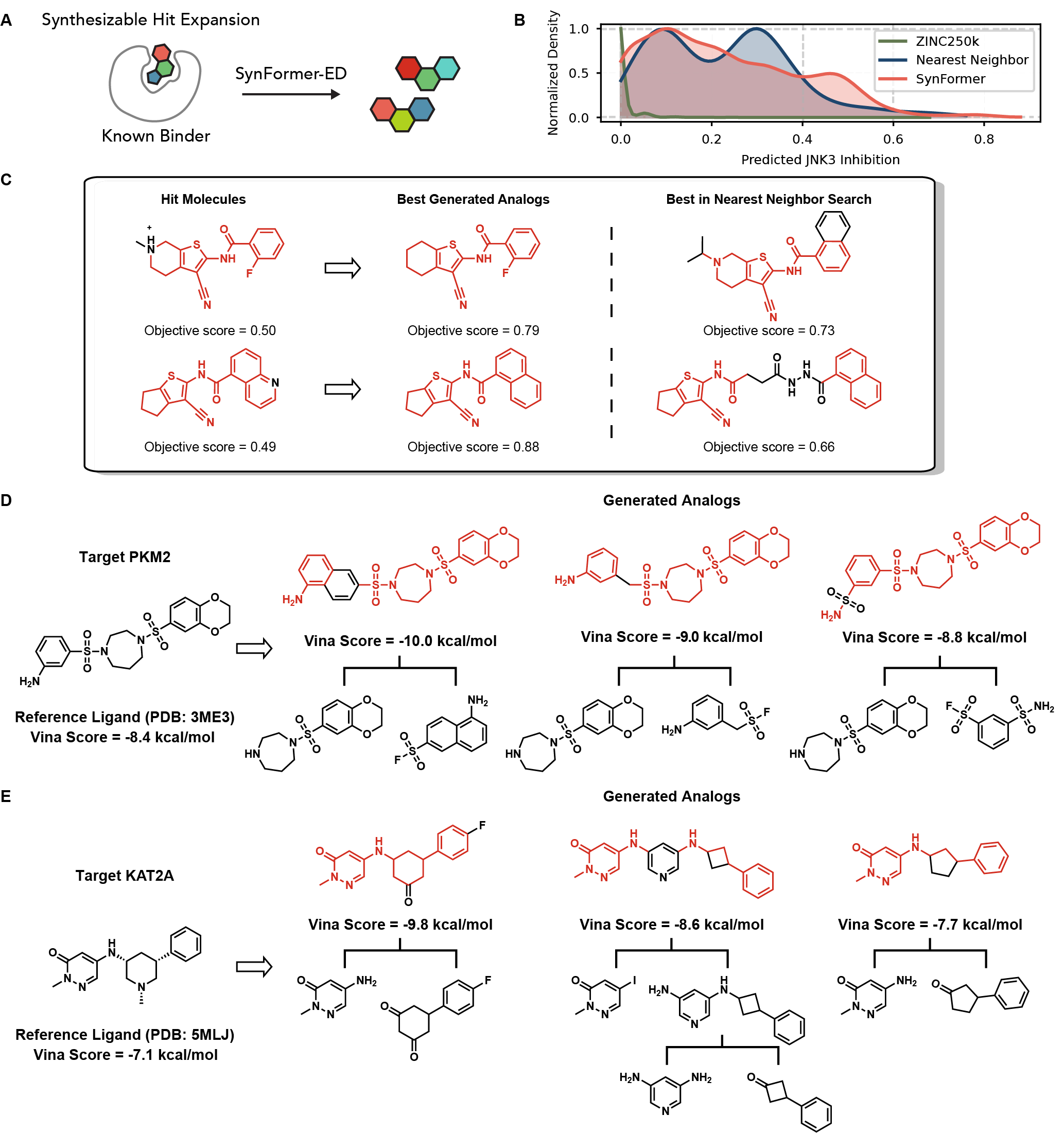}
\caption{Application of SynFormer in hit expansion. (A) Schematic illustrating SynFormer-ED expanding a known hit compound into structurally similar, synthesizable analogs. (B) Normalized distribution of predicted JNK3 inhibition scores for screening ZINC250k, nearest neighbors in Enamine REAL of the hits, and SynFormer-generated analogs of the hits, highlighting the enrichment of high-scoring ligands among SynFormer analogs. Note that the distributions are normalized to a peak value of 1 for a clearer comparison. (C) Examples of hits from screening ZINC250k and corresponding best-scored generated analogs, alongside the best-scored molecules identified in the nearest neighbor search within Enamine REAL, demonstrate SynFormer-ED’s ability to generate high scoring, synthesizable compounds. The motifs retained from the hits are colored in red. (D and E) Expanding experimentally validated ligands for the PKM2 target (PDB: 3ME3) (D) and the KAT2A target (PDB: 5MLJ) (E). Three representative analogs are shown for each target, along with their synthetic pathways. The structural motifs retained from the hits are colored in red. }\label{fig:hit_expansion}
\end{figure*}

Building on SynFormer-ED’s ability to explore local synthesizable chemical space around reference molecules, we applied it to hit expansion within synthesizable chemical space. By allowing the model to select sub-optimal intermediate choices of building blocks or reactions via a beam search, SynFormer-ED can generate sets of close analogs around known hits, as illustrated in \ref{fig:hit_expansion}A. This approach aims to produce additional molecules that preserve the core motifs and overall structural integrity, potentially enhancing design objectives and expanding the pool of candidates for downstream processes, such as hit-to-lead optimization.

Following the setup in \cite{levin2023computer}, we screened ZINC250k \cite{kusner2017grammar}, a subset of ZINC database \cite{sterling2015zinc}, using a predictive model for inhibition activity against c-Jun NH2-terminal kinase 3 (JNK3) \cite{kuan2003critical,sun2017excape,li2018multi}. The top 10 scoring molecules, with scores ranging from 0.49 to 0.68 (the higher, the better), were selected as hits and input into SynFormer-ED to generate analogs, which were then evaluated with the JNK3 inhibition predictor. As a baseline, a nearest neighbor search within Enamine REAL (approximately 7 billion molecules) was performed for the hits, retrieving approximately 100 of the most similar molecules for each hit and evaluating them for JNK3 inhibition. While both SynFormer-ED and the nearest neighbor search identified additional molecules with scores higher than those from a simple screening of ZINC250k, SynFormer-generated analogs exhibited a larger enrichment in the high-scoring region (Figure \ref{fig:hit_expansion}B). We present two sets of examples in Figure \ref{fig:hit_expansion}C. Though both methods produced analogs retaining the overall structure with higher predicted inhibition than the original hits, the SynFormer-ED analogs consistently achieved higher scores compared to those from the nearest neighbor search.

To further validate SynFormer-ED's capability in a more realistic hit expansion scenario, we used experimentally verified ligands for Human Pyruvate Kinase M2 (PKM2, PDB ID: 3ME3) \cite{anastasiou2012pyruvate} and Lysine Acetyltransferase 2A (KAT2A, PDB ID: 5MLJ) \cite{humphreys2017discovery} as reference molecules and evaluated their Vina docking scores. SynFormer-ED was used to expand the reported ligands, generating 191 analogs for PKM2, with 179 of them with Tanimoto similarities greater than 0.5 and Vina score changes lower than 1 kcal/mol. For KAT2A, SynFormer-ED generated 200 analogs, 111 of which met the criteria of Tanimoto similarity above 0.5 and Vina score changes lower than 1 kcal/mol. Figure \ref{fig:hit_expansion}D and E present three generated analogs for ligands targeting PKM2 and KAT2A, respectively, showing competitive Vina scores while preserving the overall structure, along with their synthetic pathways (see Figure S11 and 12 in the Supporting Information for more examples). Notably, in the case of KAT2A, the synthetic pathways of the generated analogs are diverse despite the structural similarity of the final analogs, highlighting that the SynFormer-based approach is not limited by the synthetic pathways of the input hits. This flexibility allows the generation of analogs with distinct synthetic routes, which traditional path-based enumeration strategies \cite{dolfus2022synthesis,levin2023computer} cannot achieve. Combined with these results, SynFormer-ED demonstrates significant potential in hit expansion, enabling the generation of practical, synthesizable analogs suitable for further drug development stages.

\subsection{Global chemical space navigation with SynFormer}
\label{sec:global}

In this section, we demonstrate the application of SynFormer to global chemical space navigation, enabling the optimization of molecular properties treated as a black-box function across broad chemical spaces \cite{gao2022sample}. This process, also known as \textit{de novo} molecular optimization, is key to exploring novel molecular designs. We first show that SynFormer-D can be fine-tuned using reinforcement learning to generate high-scoring molecules (Figure \ref{fig:global_exploration}A). Furthermore, we demonstrate how SynFormer-ED can be integrated as a mutation step within an evolutionary algorithm (Figure \ref{fig:global_exploration}C), achieving state-of-the-art sample efficiency while constraining the design space to synthesizable chemical space.

\subsubsection{Molecular optimization by fine-tuning SynFormer-D with reinforcement learning}
\label{sec:sfrl}

\begin{figure*}[t!]
\centering
\includegraphics[width=0.85\linewidth]{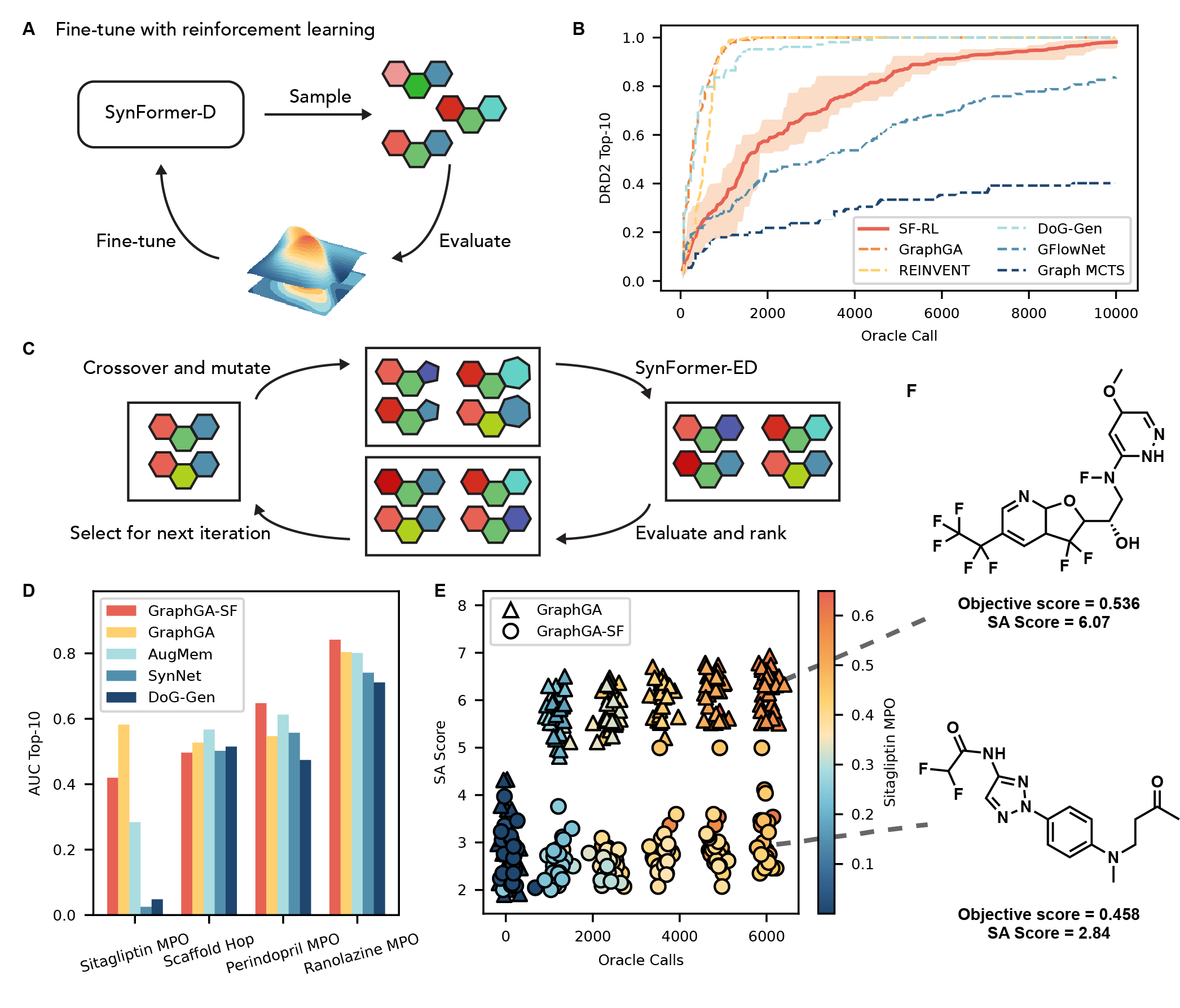}
\caption{Application of SynFormer in global chemical space exploration. (A) Illustration of fine-tuning SynFormer-D with reinforcement learning. (B) Performance comparison of SynFormer fine-tuning with reinforcement learning (SF-RL) against other popular methods, showing the average top-10 molecule scores versus the number of oracle calls \cite{gao2022sample}. The plots represent the mean performance curves of 5 independent runs, with the shaded region for SF-RL indicating the range across these runs. (C) Illustration of a genetic algorithm with SynFormer-ED used for mutation steps. (D) AUC Top-10 performance comparison across different molecular design methods (GraphGA-SF, GraphGA \cite{jensen2019graph}, AugMem \cite{guo2024augmented}, SynNet \cite{gao2021amortized}, DoG-Gen \cite{bradshaw2020barking}) for four tasks from GuacaMol \cite{brown2019guacamol}. (E) Distribution of SA Scores \cite{ertl2009estimation} for the top 25 molecules at various optimization steps, with colors representing objective scores, demonstrating how SynFormer effectively constraints its design space to synthesizable space exclusively. (F) The best molecules generated by GraphGA (top) and GraphGA-SF (bottom) show that GraphGA-SF identifies a more synthetically tractable candidate with a reduced SA score, albeit with a minor sacrifice in the objective score.}\label{fig:global_exploration}
\end{figure*}

We fine-tuned SynFormer-D using reinforcement learning (RL) to guide its generation toward high-scoring molecules. Specifically, we adopted a variant of the REINFORCE algorithm \cite{williams1992simple}, which iteratively generates a batch of molecules, evaluates their properties, and fine-tunes the model parameters with the goal of maximizing the desired properties (detailed methods can be found in the Methods section). In our experiments, we optimized predicted binding affinity against the dopamine receptor D$_2$ (DRD2) \cite{vallar1989mechanisms} as an exemplary oracle function. As shown in Figure \ref{fig:global_exploration}B, our method (denoted as SF-RL) successfully biases generation toward high-scoring molecules and outperforms several popular optimization methods \cite{bengio2021flow, jensen2019graph}. Although SF-RL’s sample efficiency is lower than the most efficient algorithms like GraphGA \cite{jensen2019graph} and REINVENT \cite{blaschke2020reinvent}, these results confirm the feasibility and effectiveness of fine-tuning SynFormer-D for \textit{de novo} molecular optimization. Given that much of the algorithm's design space of RL is yet to be explored, the potential for additional algorithmic improvements in RL-based fine-tuning remains substantial.

\subsubsection{Molecular optimization by using SynFormer-ED as a mutation operator within a genetic algorithm}
\label{sec:gasf}

Lastly, we demonstrate a genetic algorithm framework incorporating SynFormer-ED projection as a mutation step. After the standard processes of crossover and mutation to diversify the candidate pool, SynFormer-ED projects all candidates into synthesizable chemical space. This step not only corrects unsynthesizable fragments and linkages, ensuring synthetic accessibility, but also introduces additional diversity into the candidate pool. Generated molecules are then scored and ranked, with the top-performing ones selected for further iterations (Figure \ref{fig:global_exploration}C). For crossover and mutation rules, we follow the GraphGA algorithm \cite{jensen2019graph}, a well-established molecular optimization method that regularly performs well in molecular optimization benchmarks \cite{gao2022sample,tripp2023genetic} and real-world applications \cite{seumer2023computational}.

We evaluated the performance of this genetic algorithm, denoted as GraphGA-SF, across four tasks from the GuacaMol benchmark \cite{brown2019guacamol}: Sitagliptin MPO, Scaffold Hop, Perindopril MPO, and Ranolazine MPO. As shown in Figure \ref{fig:global_exploration}D, GraphGA-SF demonstrated optimization efficiency comparable to the original GraphGA and the state-of-the-art RL method, Augmented Memory (denoted as AugMem) \cite{guo2024augmented}, while significantly outperforming previous methods for synthesizable molecular optimization. Notably, in the Sitagliptin MPO task, where the primary objective score often favored unsynthesizable molecules, conventional methods without synthesizability constraints tended to generate exclusively unsynthesizable compounds (See SA Score in Figure \ref{fig:global_exploration}E and structure in Figure \ref{fig:global_exploration}F). On the other hand, synthesis-centric design methods, such as SynNet \cite{gao2021amortized} and DoG-Gen \cite{bradshaw2020barking}, struggled to find meaningful optimization signals due to the sparse signal and their limited coverage of chemical space. In contrast, GraphGA-SF matched the performance of GraphGA while ensuring each design has a synthetic pathway like other synthesis-centric methods. We highlight the best designs from each genetic algorithm in Figure \ref{fig:global_exploration}F, showing that the molecules generated by GraphGA-SF were significantly more synthesizable, with only a modest sacrifice in the main objective score. 

Our results demonstrate that by combining the optimization capability of GraphGA with SynFormer-ED’s ability to restrict the design space to synthesizable compounds, GraphGA-SF can effectively and efficiently optimize molecular properties while maintaining synthetic feasibility. Furthermore, SynFormer-ED’s integration as a mutation operator exemplifies its potential as a generalizable component in molecular design algorithms, allowing these frameworks to confine their search to synthesizable chemical spaces. This adaptability enables SynFormer-ED to leverage ongoing advances in \textit{de novo} molecular design algorithms, offering a versatile and robust solution to the problem of synthesizable molecular design.

\section{Discussion}
\label{sec:discussion}

In this paper, we introduced SynFormer, a generative modeling framework designed to efficiently explore and navigate synthesizable chemical space. We curated a set of reaction rules that allow the model to construct a synthesizable chemical space beyond Enamine REAL Space, ensuring that the designed molecules are feasible for synthesis subject to the reliability of such rules. By integrating transformer architectures with a diffusion module, SynFormer enhances the ability to make beneficial choices of building blocks and reactions during autoregressive bottom-up pathway generation, surpassing previous models for synthesizable molecular design. We demonstrated two key applications of SynFormer: local chemical space exploration and global optimization within synthesizable chemical space, both of which yielded successful results. Additionally, we showed that SynFormer is a scalable architecture, with the potential for further performance improvements with increased resources. While our focus was on applications in drug discovery, the models are readily applicable to other domains of small organic molecule design \cite{yu2020molecular,diederichsen2022electrochemical}.

There are several areas where we expect further improvements can be made. First, the reconstruction rate of SynFormer-ED is not perfect, suggesting that certain regions of synthesizable chemical space may remain inaccessible, even though the model possess the reaction templates and building blocks that should make it possible to reach them. Additionally, RL-based fine-tuning in SynFormer-D currently requires a relatively large number of oracle calls to achieve competitive results. Another limitation lies in the quality and coverage of reaction templates and building blocks. These templates do not account for stereochemistry, meaning that chiral centers can only be inherited from building blocks, not installed during synthesis. Furthermore, the Enamine building blocks we choose are designed for combinatorial library construction and do not comprehensively cover commercially available materials. These constraints may limit the model's applicability to somewhat flat, linearly-constructed sp\textsuperscript{2}-rich structures and restrict its use for more complex molecules like natural products \cite{mullowney2023artificial}. Expanding the SynFormer framework to include more sophisticated reactions and a broader range of building blocks is possible, but will require further validation.

\section{Methods}
\label{sec:method}

\subsection*{Synthesizable Chemical Space Construction}

A synthesizable chemical space is defined as a set of molecules that can be accessed from purchasable starting materials (building blocks) through a series of reliable chemical transformations. One well-known example of such a chemical space is Enamine REAL Space (REadily AccessibLe) \cite{grygorenko2020generating}, which consists of 48 billion molecules. These molecules can be synthesized by applying 1-3-step synthetic transformations derived from approximately 170 well-validated parallel synthesis protocols to over 143,000 qualified building blocks. The synthesis and delivery of molecules within the REAL Space typically takes 3-4 weeks, with an average synthesis success rate of 80\%, which attests to the validity of the way of constructing synthesizable chemical space. In this work, we aim to cover the REAL Space and extend it further by incorporating additional reactions and a greater number of reaction steps to mimic a general organic synthesis endeavor.

\subsubsection*{Reaction templates}
Reaction rules are encoded as strings using the SMARTS grammar \cite{daylight_smarts}, known as reaction templates. We created a set of 86 reaction templates, which directly implement the 169 reactions used in the January 2022 version of the REAL Space. This set includes both bi-molecular and tri-molecular reactions, with deprotection reactions treated as separate entities. Reactions that can be represented by the same transformation rule are encoded using a single SMARTS string. The combination of this reaction template set and the Enamine building block list should cover a space similar to the REAL Space.

In addition to linking building blocks, we also considered more general chemical synthesis processes. We manually selected 29 additional reaction templates from Hartenfeller et al.'s \cite{hartenfeller2011collection} and Button et al.'s \cite{button2019automated} template collection, which were not covered by the initial 86 reactions. These reactions include well-known organic transformations such as the Wittig reaction \cite{wittig1953reaktionsweise}, Buchwald-Hartwig reaction \cite{hartwig1998transition}, and various functional group conversions. This expanded set of reactions complements the REAL reactions by providing a closer approximation to general laboratory organic synthesis.

We acknowledge certain limitations in our templates: no reaction conditions, regioselectivity, or functional group compatibility are currently considered. The templates represent potential transformations, meaning that the chemical space we constructed likely represents a superset of the actual synthesizable chemical space. Furthermore, stereochemistry is not considered in our reaction rules, so stereocenters can only be inherited from the building blocks and cannot be installed during synthesis.

\subsubsection*{Building blocks}
The building block list comprises reagents or intermediates that are commercially available and can be directly purchased from the market. This list can include any set of purchasable molecules, including those found in Enamine REAL Space. For this study, we utilized Enamine's U.S. in-stock collection of building blocks, which was available for delivery as of October 1st, 2023. The building block list contains a total of 223,244 molecules. This selection was made to ensure the accessibility and immediate availability of the building blocks for synthesis.

\subsubsection*{Representing synthetic pathways using postfix notation}

In SynFormer, synthetic pathways are represented using the \textit{postfix notation of synthesis} \cite{luo2024projecting}. In this format, the operators (chemical reactions) follow their operands (molecular building blocks), providing a structured and concise way to represent reaction sequences. Each synthesis pathway is encoded as a sequence of tokens, where each token represents either a reaction step or a building block. This linear sequence simplifies the representation of converging synthetic pathways by eliminating the need for ``parentheses'' to indicate the order of operations, making it particularly amenable to autoregressive modeling. This approach enables efficient handling of complex synthetic routes while ensuring that each step in the pathway is captured in the correct order.

\subsection*{SynFormer}

The SynFormer framework is a generative modeling architecture designed to explore synthesizable chemical space by generating synthetic pathways for molecules rather than molecular graphs. SynFormer is based on a transformer architecture with a diffusion module to select molecular building blocks. We designed two variants of the model to handle distinct molecular design tasks:
\begin{itemize}
    \item SynFormer-D, a decoder-only model designed for generating new molecules or optimizing molecules based on feedback from a black-box objective function. 
    \item SynFormer-ED, an encoder-decoder model that generates synthetic pathways based on a given reference molecule, enabling local chemical space exploration. 
\end{itemize}

\subsubsection*{Transformer decoder}
SynFormer adopts the transformer decoder architecture as its backbone \cite{vaswani2017attention}.
The decoder generates postfix notations of synthesis in an auto-regressive manner, which is, predicting the next token according to previous tokens.
We assign embedding vectors to each reaction template and the [START] token, and convert building blocks fingerprint into continuous-valued vectors with a multi-layer perceptron (MLP) in the beginning.
To indicate the position of each token within the sequence, we add positional encodings \cite{vaswani2017attention} to the embedding vectors.

The transformer decoder gathers information for each token from its preceding tokens via the multi-head attention mechanism \cite{vaswani2017attention}, where each token's embedding vector is first projected to query, key, and value vectors, and then updated by aggregating the values of other tokens weighted by the softmax-ed inner product of the query vector and the other token's key vector.

Following the transformer decoder, we use a classifier based on an MLP to predict the next token's type.
If it is predicted as an [END] token, the decoding process will be terminated.
If it is predicted as a reaction, another classifier is used to predict the probability of each reaction template being the next token.
When the next token is predicted as a building block, a diffusion module is used to generate molecular fingerprints which are subsequently used to retrieve building blocks from the database.

\subsubsection*{Transformer encoder}
The transformer decoder can be conditioned by a transformer encoder, i.e., given reference molecule represented by a SMILES string.
The transformer encoder produces vector representations for each token in the SMILES string, which the decoder can attend to via the cross attention mechanism \cite{vaswani2017attention}.

\subsubsection*{Diffusion module for building block fingerprints}

We build a denoising diffusion model to learn the distribution of molecular fingerprints, represented as $n$-dimensional binary vectors, where $n$ denotes the number of bits in the fingerprint. This distribution is modeled as a joint Bernoulli binary distribution of $n$ dimensions.

During training, the forward diffusion process perturbs the fingerprint vector by randomly flipping each bit according to the following noise distribution:
\begin{equation}
q(x_t^{(j)} | x_0^{(j)}) = \operatorname{Bernoulli}(x_t^{(j)} \mid \bar{\alpha_t} x_0^{(j)} + (1 - \bar{\alpha_t}) \cdot \frac{1}{2} ),
\end{equation}
where $x_0^{(j)}$ represents the $j$-th bit of the ground truth fingerprint vector, and $x_t^{(j)}$ represents the perturbed bit.
Here, $\bar{\alpha_t}$ is a value that controls the noise level, which we define using a formulation similar to multinomial diffusion \cite{hoogeboom2021argmax}.
$\bar{\alpha_t}$ starts at 1 and monotonically decreases to 0.
When $\bar{\alpha_t} = 0$, the distribution $q$ becomes the uniform Bernoulli distribution, where the noise level reaches the maximum.
The denoiser network learns the bitwise Bernoulli distribution of ground truth fingerprints conditioned on the perturb fingerprint and the embeddings from the transformer decoder.
It is trained by minimizing the binary cross entropy loss that measures the discrepancy between the distribution and the ground truth fingerprint.

The inference process is known as the reverse diffusion process, where we start with a random bit vector drawn from the uniform Bernoulli distribution, and denoise it iteratively with the denoiser network.
Finally, the denoised bit vector is used as a fingerprint to retrieve building block molecules from the database.

\subsubsection*{Reinforcement learning fine-tuning}

For global optimization tasks, SynFormer-D was fine-tuned using a variant of the REINVENT algorithm \cite{blaschke2020reinvent} based on REINFORCE \cite{williams1992simple}, a type of reinforcement learning (RL). Our RL setup was designed to minimize the following loss function, which incentivizes the model to bring the pseudo-log-likelihood (PLL) of generating a synthetic pathway closer to the objective score of the product molecules:
\begin{equation}
    L (\theta) = \mathbb{E} \left[ \left( \text{PLL}_{\theta} (p, m) - \sigma \cdot S (m) \right) ^2 \right],
\end{equation}
where $\theta$ represents the trainable network parameters, $p$ is the synthetic pathway, $m$ is the product molecule, $S(m)$ is the objective score of molecule $m$, typically scaled between 0 and 1 (the higher, the better), $\sigma$ is a constant scaling factor, $\text{PLL}_{\theta}(p, m)$ is the pseudo-log-likelihood of generating the pathway $p$ leading to molecule $m$. The pseudo-log-likelihood, $\text{PLL}_{\theta}(p, m)$, is computed as the sum of three terms: the log-likelihood of selecting each token (reaction step or building block), the log-likelihood of selecting each reaction type, and the empirical log-likelihood value of the building block fingerprints:
\begin{equation}
    \text{PLL}_{\theta}(p) = \sum_{i} \log P_{\theta}(x_i) + \log \hat P_{\theta}(x_{\text{building block}}),
\end{equation}
where $x_i$ represents each component of a synthetic path (token, reaction). The empirical likelihood of the building block fingerprints is approximated by:
\begin{align}
    \hat P_{\theta}(x_{\text{building block}}) &= x_{\text{denoised}} \cdot x_{\text{target}} \nonumber \\
    &\quad + (1 - x_{\text{denoised}}) \cdot (1 - x_{\text{target}}),
\end{align}
where $x_{\text{denoised}}$ is the denoised fingerprint vector, $x_{\text{target}}$ is the ground truth fingerprint. 


\subsubsection*{Genetic algorithm with SynFormer-ED as a mutation operator}

We integrated SynFormer-ED as a mutation step into an evolutionary framework for global chemical space exploration and optimization. The genetic algorithm follows a standard workflow: starting with a pool of candidates randomly sampled from the ZINC database, pairs of molecules are selected to undergo crossover, producing an offspring pool. Mutation is then applied to each offspring with a specific probability, followed by an additional mutation step where SynFormer-ED projects all offspring into synthesizable space. This step ensures that unsynthesizable fragments are corrected and adds additional diversity to the candidate pool. The offspring pool will be evaluated, and the highest-scored ones will form the next generation for the next iteration. The crossover and mutation rules were adopted from the GraphGA \cite{jensen2019graph} algorithm; specifically, crossover involves graph matching and exchanging molecular halves, while mutation employs a set of hand-coded rules including both atom- and fragment-level modifications. We adopted the hyperparameters from \cite{tripp2023genetic}. 


\subsection*{Evaluation Details}
This section describes the precise settings used in the evaluation of SynFormer-ED and SynFormer-D.

\subsubsection*{Reconstruction and local chemical space exploration}

SynFormer-ED was employed for reconstruction, analog generation, and hit expansion. The workflows for these tasks are similar: SynFormer-ED encodes the input molecule and decodes it into synthetic pathways, using a search width of 24 and an exhaustiveness of 64 during decoding. For reconstruction and analog generation, Tanimoto (Jaccard) similarity \cite{jaccard1901etude} to the input molecule based on Morgan fingerprints \cite{rogers2010extended} was evaluated for each generated molecule, with the most similar molecule selected as the output. In hit expansion, generated molecules were evaluated based on both the design objective score and Tanimoto similarity.

We tested the reconstruction performance on a random sample of 1,000 molecules from Enamine's REAL Space \cite{grygorenko2020generating} and the ChEMBL database \cite{zdrazil2024chembl}. Molecules were sampled from the REAL Diversity Set, which includes 48.2 million molecules and maximizes diversity to represent the broader REAL Space. These molecules comply with the Rule of 5 (Ro5) \cite{lipinski1997experimental} and Veber criteria \cite{veber2002molecular}, making them suitable for drug-like properties. ChEMBL samples were taken from version 29 of the database, released in 2021.

The Synthetic Accessibility (SA) Score \cite{ertl2009estimation}, a heuristic measure evaluating structural complexity based on fragment frequency in PubChem \cite{kim2016pubchem}, was used to assess the synthesizability of the generated molecules. For synthesizable analog generation, 10 objective scores from the GuacaMol benchmark \cite{brown2019guacamol} were used, which involve optimizing various physicochemical properties, similarities or dissimilarities to known drugs, and specific structural motifs.

In hit expansion, we also used a predictive model \cite{li2018multi} to assess inhibition against c-Jun N-terminal Kinase-3 (JNK3), a member of the mitogen-activated protein kinase family. This model is a random forest classifier utilizing ECFP6 fingerprints trained on the ExCAPE-DB dataset \cite{sun2017excape}.

AutoDock Vina \cite{trott2010autodock} was employed to evaluate binding affinities against protein targets in both structure-based drug design and hit expansion tasks. Protein structures and pocket information were sourced from the LIT-PCBA dataset \cite{tran2020lit}, and an exhaustiveness setting of 16 was used for most calculations.

For baseline comparisons in analog finding, we conducted a nearest neighbor search within Enamine REAL. This was done by accessing the Enamine website (https://new.enaminestore.com/draw-search) and downloading approximately 100 nearest neighbors by adjusting the similarity threshold value.

\subsubsection*{Global chemical space exploration}

For both SF-RL and GraphGA-SF, we employed these models to solve molecular optimization problems without predefined starting points \cite{gao2022sample}. In each experiment, we assumed the presence of an oracle, i.e., a function that evaluates the desired property of a molecule, providing the ground truth value. As property evaluation is often time- and resource-intensive in real-world scenarios, each experiment was limited to 10,000 oracle calls, meaning up to 10,000 molecules could be evaluated. During optimization, all evaluated molecules were recorded, and the primary performance metric was the area under the curve (AUC) of the top-$K$ average property value versus the number of oracle calls (\textit{AUC top-$K$}). The reported AUC values were min-max scaled to the range [0, 1].

We evaluated the global chemical space optimization capabilities of SF-RL and GraphGA-SF on optimizing DRD2 inhibition \cite{olivecrona2017molecular} and four multi-objective properties from the GuacaMol benchmark \cite{brown2019guacamol}: Sitagliptin MPO, Scaffold Hop, Perindopril MPO, and Ranolazine MPO. DRD2 inhibition was assessed using a classifier trained on the ExCAPE-DB dataset \cite{sun2017excape}, employing a support vector machine (SVM) with a Gaussian kernel and ECFP6 fingerprints to distinguish active from inactive compounds for the Dopamine Receptor D2 (DRD2) \cite{olivecrona2017molecular}. The multi-objective tasks from the GuacaMol benchmark are described above \cite{brown2019guacamol}. All oracle functions were accessed through the Therapeutic Data Commons (TDC) \cite{huang2021therapeutics,huang2022artificial}.

The baseline methods compared in this study include:
\begin{itemize}
    \item \textbf{GraphGA} \cite{jensen2019graph} is a popular heuristic algorithm inspired by natural evolutionary processes, featuring crossover rules derived from graph matching and mutations applied at both the atom- and fragment-level.
    \item \textbf{Graph MCTS} \cite{jensen2019graph} uses Monte Carlo Tree Search to explore molecular structures by locally searching each branch of the current state (molecule or partial molecule) and selecting the most promising candidates based on property scores for subsequent iterations.
    \item \textbf{REINVENT} \cite{blaschke2020reinvent} utilizes a policy-based reinforcement learning (RL) approach where agents take actions in an environment to maximize cumulative rewards, tuning recurrent neural networks (RNNs) to generate SMILES strings.
    \item \textbf{Augmented Memory (AugMem)} \cite{guo2024augmented} integrates SMILES-based reinforcement learning with augmented training through experience replay and selective memory purge to prevent model collapse. This approach achieves state-of-the-art sample efficiency, demonstrating the highest reported performance in the PMO benchmark.
    \item \textbf{GFlowNet} \cite{bengio2021flow} is a generative AI model that treats the generative process as a flow network and trains it with a temporal difference-like loss function, aligning the generation process with the target property distribution by matching the property of interest to the flow volume.
    \item \textbf{DoG-Gen} \cite{bradshaw2020barking} models synthetic pathways as Directed Acyclic Graphs (DAGs) and uses an RNN-based generator, optimizing via an iterative learning method that incorporates high-scoring molecules into the training data for successive fine-tuning of the generative model.
    \item \textbf{SynNet} \cite{gao2021amortized} is a synthesis-based method that applies a genetic algorithm to molecular fingerprints, decoding them into synthetic pathways using trained neural networks.
\end{itemize}
The implementations of these methods were adapted from their original publications with minor modifications, and hyperparameters were set according to the specifications in \cite{gao2022sample}, except AugMem which is from the original publication.

\section*{Acknowledgments}
This research was supported by the Office of Naval Research under grant number N00014-21-1-2195 and the AI2050 program at Schmidt Futures under grant number G-22-64475. Any opinions, findings, and conclusions or recommendations expressed in this material are those of the author(s) and do not necessarily reflect the views of the Office of Naval Research. W.G. received additional funding from the Google Ph.D. fellowship. We thank Kaiming He and Yurii Moroz for their helpful discussions and comments on the manuscript.

\section*{Reproducibility Statement}
All code and releasable data can be found at \url{https://github.com/wenhao-gao/synformer}, including instructions in a README file. The model weights can be accessed on \url{https://huggingface.co/whgao/synformer}. Appendix \ref{sec:details} provides more details about the experimental setup and implementation details.

\bibliographystyle{unsrt}  
\bibliography{references}

\appendix
\newpage

\section{Additional method details}
\label{sec:details}

\subsection{Model architectures}

\subsubsection{Transformer decoder}
SynFormer adopts the standard Transformer decoder architecture \cite{vaswani2017attention} to generate postfix notation of synthesis in an auto-regressive manner.
A postfix notation of synthesis may contain four types of tokens: a [START] token, [END] token, reaction token, and building block token.
Each token, regardless of its type, is represented as a 768-dimensional embedding vector that is differentiable during training.
Specifically, the embedding vectors of start token, end token, and reaction tokens are stored in a look-up table implemented by PyTorch's \texttt{nn.Embedding} layer.
Note that, as we use a fixed set of reactions consisting of only 169 reaction templates, it is practical to store the embedding vectors for each of them.
However, the building block database, containing more than 200k molecules, is too large to assign embedding vectors to each building block separately.
Thus, we use a multi-layer perceptron (MLP), to convert 2048-bit Morgan fingerprints of building blocks into 768-dimensional continuous vectors.
This MLP consists of three layers, with hidden dimensions of 1,536 and ReLU activation. 

To inform the model with the location of each token, we add sinusoidal positional encoding \cite{vaswani2017attention} to embedding vectors:
\begin{equation}
\label{eq:posenc}
    \mathbf{p}_i = \begin{cases}
        \sin\left( \frac{j}{10000^\frac{2i}{d}} \right), & i = 0, 2, \ldots ,766 \\
        \cos\left( \frac{j}{10000^\frac{2i}{d}} \right), & i = 1, 3, \ldots , 767 \\
    \end{cases} \ ,
\end{equation}
where $i$ denotes the dimension index, $d=768$ denotes the embedding dimension, and $j$ denotes the token position within the sequence.

The Transformer decoder network operates on the token embedding vectors.
It consists of 10 attention layers with residual connection and layer normalization in between.
In SynFormer-ED, the attention layer contains both self-attention and cross-attention that gathers information from the encoder, while in SynFormer-D, only self-attention is used.
Each attention layer contains 16 attention heads. Each head projects the feature vectors into 256-dimensional query, key, and value vectors, which are then transformed into output vectors via the attention mechanism.
Output vectors from different attention heads are concatenated into a 2,048-dimensional vector, which is then mapped into a 768-dimensional vector used to update the token embedding via residual connection and layer normalization.

\subsubsection{Transformer encoder}
SynFormer-ED contains a Transformer-based encoder network that conditions the decoder with a given SMILES string.
The input SMILES string is split in to a list of tokens according to the following vocabulary:

\texttt{H, He, Li, Be, B, C, N, O, F, Ne, Na, Mg, Al, Si, P, S, Cl, Ar, K, Ca, Sc, Ti, V, Cr, Mn, Fe, Co, Ni, Cu, Zn, Ga, Ge, As, Se, Br, Kr, Rb, Sr, Y, Zr, Nb, Mo, Tc, Ru, Rh, Pd, Ag, Cd, In, Sn, Sb, Te, I, Xe, Cs, Ba, La, Ce, Pr, Nd, Pm, Sm, Eu, Gd, Tb, Dy, Ho, Er, Tm, Yb, Lu, Hf, Ta, W, Re, Os, Ir, Pt, Au, Hg, Tl, Pb, Bi, Po, At, Rn, Fr, Ra, Ac, Th, Pa, U, Np, Pu, Am, Cm, Bk, Cf, Es, Fm, Md, No, Lr, Rf, Db, Sg, Bh, Hs, Mt, Ds, Rg, Cn, Nh, Fl, Mc, Lv, Ts, Og, b, c, n, o, s, p, 0, 1, 2, 3, 4, 5, 6, 7, 8, 9, [, ], (, ), ., =, \#, -, +, $\backslash$, /, :, ~, @, ?, >, *, \$, \%}

Each token in the list is represented by a 768-dimensional embedding vector, and position encodings (Eq. \ref{eq:posenc}) are added to the embedding vector.
The Transformer encoder contains 6 layers, each of which has 16 attention heads with 256-dimensional query, key, and value vectors.

\subsubsection{Scaling of model size}
Table~\ref{tab:scaling} below shows the hyper-parameters for SynFormer-ED of different sizes measured by the number of trainable parameters.
These hyper-parameter sets are used to investigate the scaling of model performance. For the remaining experiments, we use the model with 229M parameters.

\begin{table}[h!]
\centering
\caption{Specific model hyper-parameters used in the scaling experiments.}
\label{tab:scaling}
\begin{tabular}{l cccc}
\toprule
Model Size & \textbf{229M} & 68M & 17.5M & 3.7M \\
\midrule
\# Feature Dimensions & 768 & 512 & 256 & 128 \\
\# Encoder Layers & 8 & 6 & 6 & 4 \\
\# Encoder Attn. Heads & 16 & 16 & 8 & 4 \\
\# Encoder Feed-forward Dimensions & 4096 & 2048 & 1024 & 512\\
\# Decoder Layers & 10 & 8 & 8 & 6 \\
\# Decoder Attn. Heads & 16 & 16 & 8 & 4 \\
\# Decoder Feed-forward Dimensions & 4096 & 2048 & 1024 & 512 \\
\bottomrule
\end{tabular}
\end{table}

\subsubsection{Diffusion fingerprint head}

The fingerprint of a building block token are modeled by an $n$-dimensional Bernoulli distribution. We use the denoising diffusion probabilistic model \cite{ho2020denoising,nichol2021improved} to model this distribution.
The denoising diffusion probabilistic model generates data by iterative denoising an initial sample drawn from a simple prior distribution.
To train a neural network capable of denoising, noisy samples based on ground truth data must be constructed via the forward diffusion process.

In SynFormer, we define the bit-wise forward diffusion process for fingerprint $\mathbf{x} = [x^{(1)}, \ldots, x^{(n)}] \in \{ 0, 1\}^n$ as, following the formulation of multinomial diffusion \cite{hoogeboom2021argmax}:
\begin{equation}
\label{eq:diffusion_forward_kernel}
    q(\mathbf{x}_t | \mathbf{x}_{t-1}) = \prod_{j=1}^{n} q(x_t^{(j)} | x_{t-1}^{(j)}) \quad \text{where} \quad
    q(x_t^{(j)} | x_{t-1}^{(j)}) = \operatorname{Bernoulli}(x_t^{(j)} \mid (1 - \beta_t) x_{t-1}^{(j)} + \beta_t \cdot \frac{1}{2} ), \quad j = 1\ldots n.
\end{equation}
Since the forward diffusion process is Markovian, it is possible to express the bit-wise distribution of $\mathbf{x}_t$ directly with the ground truth fingerprint $\mathbf{x}_0$:
\begin{equation}
\label{eq:diffusion_forward}
    q(\mathbf{x}_t | \mathbf{x}_{0}) = \prod_{j=1}^{n} q(x_t^{(j)} | x_0^{(j)}) \quad \text{where} \quad
    q(x_t^{(j)} | x_0^{(j)}) = \operatorname{Bernoulli}(x_t^{(j)} \mid \bar{\alpha_t} x_0^{(j)} + (1 - \bar{\alpha_t}) \cdot \frac{1}{2} ), \quad j = 1\ldots n.
\end{equation}
where $\alpha_t = 1 - \beta_t$ and $\bar{\alpha_t} = \prod_{\tau=1}^t \alpha_\tau$.
This expression enables efficient one-step generation of $\mathbf{x}_t$ given $\mathbf{x}_0$ at training time, while it requires $t$ steps if we follow Eq.~\ref{eq:diffusion_forward_kernel}.
Intuitvely, $1-\bar{\alpha_t}$ determines the chance of the fingerprint bit being resampled uniformly at timestep $t$, and it should increase monotonically from $t = 0$ to $T$.
We adopt the noise schedule in terms of $\bar{\alpha_t}$ proposed by Nichol and Dhariwal \cite{nichol2021improved}:
\begin{equation}
    \bar{\alpha_t} = \frac{f(t)}{f(0)}, \ f(t) = \cos \left( \frac{\frac{t}{T} + s}{1 + s} \cdot \frac{\pi}{2} \right)^2.
\end{equation}
From this definition, $\beta_t$ and $\alpha_t$  can be obtained by $\beta_t = 1 - \frac{\bar{\alpha_t}}{\bar{\alpha_{t-1}}}$ and $\alpha_t = 1 - \beta_t$ respectively. In practice, $\beta_t$ is clipped to be no larger than 0.999 to prevent instability.

To denoise a noisy fingerprint $\mathbf{x}_t$ into the less noisy $\mathbf{x}_{t-1}$ which eventually leads to $\mathbf{x}_{0}$, the denoiser network should learn the following posterior distribution derived from Eq.\ref{eq:diffusion_forward_kernel} and Eq.\ref{eq:diffusion_forward}:
\begin{equation}
    q(\mathbf{x}_{t-1} | \mathbf{x}_t , \mathbf{x}_0) = \prod_{j=1}^{n} q(x_{t-1}^{(j)} | x_t^{(j)} , x_0^{(j)}) \quad \text{where} \quad
    q(x_{t-1}^{(j)} | x_t^{(j)} , x_0^{(j)}) = \operatorname{Bernoulli}\left( x_{t-1}^{(j)} \middle| \frac{\theta_1(x_t^{(j)} , x_0^{(j)})}{\theta_1(x_t^{(j)} , x_0^{(j)}) + \theta_0(x_t^{(j)} , x_0^{(j)})} \right),
\end{equation}
where
\begin{align}
    \theta_1(x_t^{(j)} , x_0^{(j)}) & = \left( \alpha_t x_t^{(j)} + (1 - \alpha_t)\cdot \frac{1}{2} \right) \cdot \left( \bar{\alpha}_{t-1} x_0^{(j)} + (1 - \bar{\alpha}_{t-1}) \cdot \frac{1}{2} \right) \\
    \theta_0(x_t^{(j)} , x_0^{(j)}) & = \left( \alpha_t (1 - x_t^{(j)}) + (1 - \alpha_t)\cdot \frac{1}{2} \right) \cdot \left( \bar{\alpha}_{t-1} (1-x_0^{(j)}) + (1 - \bar{\alpha}_{t-1}) \cdot \frac{1}{2} \right).
\end{align}
In this posterior distribution, $\mathbf{x}_0$ is the unknown variable to be predicted, which we parameterize with a neural network conditional on both the current noisy fingerprint $\mathbf{x}_t$ and the embedding vector from the Transformer decoder $\mathbf{h}$, leading to the following reverse diffusion process:
\begin{equation}
    p(\mathbf{x}_{t-1} | \mathbf{x}_t) = \prod_{j=1}^{n} \operatorname{Bernoulli}\left(x_{t-1}^{(j)} \middle| 
    \frac{\theta_1(x_t^{(j)} , p_0^{(j)})}{\theta_1(x_t^{(j)} , p_0^{(j)}) + \theta_0(x_t^{(j)} , p_0^{(j)})}
    \right) \quad \text{where} \quad
    p_0^{(j)} = \operatorname{sigmoid}( \operatorname{DenoiserNet}( \mathbf{x}_{t}, \mathbf{h} )[j] ).
\end{equation}

At training time, we randomly sample a step index $t$ from $\mathcal{U}\{1, T\}$, perturb the ground truth fingerprint according to Eq.~\ref{eq:diffusion_forward}, and finally compute and back-propagate through the following training loss to align the predicted distribution to the ground truth posterior:
\begin{equation}
    L_\text{fingerprint} = \mathbb{E}_{t\sim \mathcal{U}\{1, T\}, q(\mathbf{x}_t | \mathbf{x}_0)} \left[ \operatorname{KL} \left( q(\mathbf{x}_{t-1} | \mathbf{x}_t , \mathbf{x}_0), p(\mathbf{x}_{t-1} | \mathbf{x}_t) \right) \right].
\end{equation}
During inference, we start with a random bit vector $\mathbf{x}_T$ sampled from uniform Bernoulli distribution, and resample it following $p(\mathbf{x}_{t-1} | \mathbf{x}_t)$ for $T $ steps.

SynFormer's diffusion module has $T=100$ timesteps and the hyper-parameter $s$ controlling the noise schedule is set to 0.01.
The DenoiserNet is a multi-layer perceptron, with input dimension 2,816 ($\dim(\mathbf{h}) + \dim(\mathbf{x})$), hidden dimension 4096, output dimension 2,048 (fingerprint size), and ReLU activation in between.

\subsection{Training details}

\subsubsection{Training data generation}
We generate random postfix notations of synthesis using the building blocks and reaction templates from the database to train SynFormer.
To ensure the validity of generated postfix notations, we leverage the \textit{synthesis stack} data structure introduced in our previous model \cite{luo2024projecting} to store intermediate reaction products during synthesis.
When training SynFormer-ED, the generated postfix notation is used to supervised the decoder part and the SMILES string of the final product is used as the input to the encoder part. 
For SynFormer-D, only postfix notations are used to train the model.
The detail of this data generation procedure is described in Algorithm~\ref{alg:training_data}.

\begin{algorithm*}[hbt!]
\caption{Training data generation \cite{luo2024projecting}}
\label{alg:training_data}
\KwIn{$m_r$, maximum number of reactions}
\KwIn{$m_a$, maximum number of atoms in the final product molecule}
\KwIn{$\mathcal{F}: M \mapsto \{ R \}$, a function that returns available reactions for molecule $M$}
\KwIn{$\mathcal{H}: R \mapsto \{ M \}$, a hash map that stores eligible molecules for each reaction}
\KwOut{an iterator of postfix notation and product molecule tuples $(P, S.\operatorname{top}())$}

\BlankLine

\SetKwFunction{push}{Push}
\SetKwFunction{append}{Append}
\SetKwFunction{yield}{Yield}
\SetKwFunction{pop}{Pop}

$S \gets []$ \tcp*{Initialize empty stack}
$P \gets []$ \tcp*{Initialize empty postfix notation}

$B_0 \gets$ a random building block\;
\push{$B_0$ onto $S$}\;
\append{$B_0$ to $P$}\;
\yield{$P$, $S.\operatorname{top}()$}\;

\While{$P.\operatorname{countReactions}() < m_r$ \textbf{and} $S.\operatorname{top}().\operatorname{countAtoms}() < m_a$}{
    
    $R \sim \mathcal{R}(S.\operatorname{top}())$ \tcp*{Randomly choose a reaction according to the molecule on the stack top}
    
    \For{$i = 1$ \KwTo $R.\operatorname{numReactants} - S.\operatorname{length}()$}{
        $B_i \sim \mathcal{H}(R)$ subject to $B_i$ and the molecules in $S$ can react via $R$\;
        \push{$B_i$ onto $S$}\;
        \append{$B_i$ to $P$}\;
    }
    
    $\{ X_i \} \gets$ \pop{$R.\operatorname{numReactants}$ elements from $S$}\;
    $Y \gets$ \textbf{Call} $R(\{X_i\})$\;
    \push{$Y$ onto $S$}\;
    \append{$R$ to $P$}\;
    
    \If{$S.\operatorname{length}() = 1$}{
        \yield{$P$, $S.\operatorname{top}()$}\;
    }
}

\end{algorithm*}

\subsubsection{Hyper-parameters}

The SynFormer models are trained using the AdamW optimizer with a learning rate of $3 \times 10^{-4}$ and a batch size of 1,024.
The training process spans approximately 730,000 steps, equivalent to a total of 742M data points.
Validation is performed every 5,000 steps.
A ReduceLROnPlateau scheduler is applied, reducing the learning rate by 20\% if the validation loss does not improve over the last 10 validation runs.
The models are trained on 8 A100 GPUs, and the total training time is about 6 days and 1 hours.

\subsection{Inference details}

\subsubsection{Sampling algorithm}
The sampling process starts with an empty stack and an empty postfix notation.
The stack and the postfix notation are updated iteratively as described in Algorithm~\ref{alg:sampling_step}.
Specifically, the SynFormer decoder first outputs the embedding vector and predicts the type for the next token based on the current postfix notation.

If the next token is predicted to be a building block, the diffusion module will generate a set of $k_\text{fingerprint}$ fingerprints.
For each fingerprint, the building block molecule with the most similar fingerprint measured by L2 distance is retrieved from the database.
$k_\text{fingerprint}$ copies of the previous stack and postfix notation are made and each building block is added to the stack and the notation of each copy, leading to $k_\text{fingerprint}$ different next states.

If the next token is predicted to be a reaction, $k_\text{reaction}$ reactions with the highest probability are selected.
For each selected reaction, the required number of reactants are popped out of the stack. If there are too few molecules in the stack, the reaction is considered inapplicable.
Then, the reactants are input to the reaction template to predict the product. If the reaction runs successfully, the reaction product will be pushed onto the stack and the reaction itself will be appended to the postfix notation to form a next state. Otherwise, the reaction will be ignored and no next state for this reaction is produced.

Finally, if the next token is an [END] token, we will check if the stack contains exactly one molecule. If it does, the sampling process is considered successful and the only molecule on the stack is the final product.
Otherwise, more than one molecule in the stack means the synthesis process is incomplete and thus is considered failed.

\subsubsection{Branched sampling}
The sampling algorithm can lead to branched synthesis pathways at each building block and reaction step. This capability is particularly useful when a specific objective, for example similarity to a given molecule in analog generation, is set.
We leverage this capability by formulating a beam-search-like algorithm as described in Algorithm~\ref{alg:batched_sampling}.
Specifically, we maintain a pool of states and each state is a tuple of synthesis stack and postfix notation.
In the beginning, the pool contains only one empty state.
At each state, we run the sampling step (Algorithm~\ref{alg:sampling_step}) for each state in the pool, remove the existing state, and add the set of next generated states back to the pool.
If the pool size exceeds the limit, we sort the state according to the some given metric and then keep only the top-ranked states. 
If a state is marked successful, it will be moved out of the pool and added to the result buffer.

\begin{algorithm*}[hbt!]
\caption{Sampling step}
\label{alg:sampling_step}
\KwIn{$P_\text{prev}$, previous postfix notation of synthesis}
\KwIn{$S_\text{prev}$, previous synthesis stack}
\KwIn{$\mathbf{c}$, features from the encoder part of SynFormer, optional}
\KwIn{$k_\text{reaction}$, branching factor of reaction}
\KwIn{$k_\text{fingerprint}$, branching factor of building block fingerprint}


\tcp{Global variables: $\operatorname{SynFormer}$, $\operatorname{BuildingBlockDatabase}$}

\KwOut{an iterator of next states in the form of postfix notation and stack tuple $(P, S)$}

\BlankLine

$\mathbf{h} \gets \operatorname{SynFormer.decoder}( \operatorname{embed}(P_\text{prev}), \mathbf{c} )$ \tcp*{Get the embedding for the next token}
$t \gets \operatorname{SynFormer.predictNextType}(\mathbf{h})$ \tcp*{Predict the type of the next token}

\BlankLine
\If{$t = \text{BuildingBlock}$}{
    \ForEach{$\mathbf{x}$ \textbf{in} $\operatorname{SynFormer.generateFingerprints}(\mathbf{h}, k_\text{fingerprint})$}{
        $P, S \gets \operatorname{copy}(P_\text{prev}, S_\text{prev})$\;
        $X \gets \operatorname{BuildingBlockDatabase.retrieve}(\mathbf{x})$\;
        \Push{$X$ onto $S$}\;
        \Append{$X$ to $P$}\;
        \Yield{$P$, $S$}\;
    }
}
\ElseIf{$t = \text{Reaction}$}{
    \ForEach{$R$ \textbf{in} top $k_\text{reaction}$ of $\operatorname{SynFormer.predictNextReaction}(\mathbf{h})$}{
        $P, S \gets \operatorname{copy}(P_\text{prev}, S_\text{prev})$\;
        $\{ X_i \} \gets $ \Pop{$R.\operatorname{numReactants}$ elements from $S$}\;
        $Y \gets$ \textbf{Call} $R(\{X_i\})$\;
        \If{there are sufficient reactants in the stack and the reaction applied successfully}{
            \Push{$Y$ onto $S$}\;
            \Append{$R$ to $P$}\;
            \Yield{$P$, $S$}\;
        }
    }
}
\ElseIf{$t = \text{[END]}$}{
    \If{$S_\text{prev}.\operatorname{length}() = 1$}{
        \Yield{$P_\text{prev}$, $S_\text{prev}$, EndFlag}\;
    }
}
\end{algorithm*}

\begin{algorithm*}[hbt!]
\caption{Branched sampling}
\label{alg:batched_sampling}
\KwIn{$k_\text{reaction}$, branching factor of reaction}
\KwIn{$k_\text{fingerprint}$, branching factor of building block fingerprint}
\KwIn{$m$, size limit of the state pool}
\KwIn{$n$, expected number of outputs}
\KwIn{$f(\cdot)$, scoring function}


\KwData{$\mathcal{M} \gets \{ (P_0, S_0) \}$ where $P_0 = \varnothing$ and $S_0 = \varnothing$, initial state pool}
\KwData{$\mathcal{N} \gets \{ \}$, output buffer}

\While{$|\mathcal{N}| < n$}{
    \ForEach{$(P, S)$ \textbf{in} $\mathcal{M}$}{
        \Remove{$(P, S)$ from $\mathcal{M}$}\;
        \ForEach{$(P_\text{next}, S_\text{next}, \text{hasEndFlag})$ \textbf{in} $\operatorname{SamplingStep}(P, S, \mathbf{c}, k_\text{reaction}, k_\text{fingerprint})$}{
            \If{hasEndFlag}{
                \Add{$(P_\text{next}, S_\text{next})$ to $\mathcal{N}$} \tcp*{Save to output buffer}
            }
            \Else{
                \Add{$(P_\text{next}, S_\text{next})$ to $\mathcal{M}$} \tcp*{Add to the state pool}
            }
        }
    }
    \Sort{$\mathcal{M}$ according to the scoring function $f(S)$}\;
    $\mathcal{M} \gets \mathcal{M}[1\ldots m]$\;
}
\Return{$\mathcal{N}$}

\end{algorithm*}

\clearpage

\section{Additional results}

\subsection{Impact of enlarged fingerprint length on performance}
The length of Morgan fingerprints is a critical parameter that determines the resolution of molecular fingerprinting. Shorter fingerprint lengths can lead to bit collisions, where two similar molecules become indistinguishable due to the limited bit space \cite{landrum2024colliding}. However, directly predicting larger fingerprint lengths has proven challenging to scale (see Figure 14 in the SI of \cite{gao2021amortized}), often necessitating a compromise at shorter lengths, such as 256 bits.

\begin{figure}[H]
\centering
\includegraphics[width=0.75\textwidth]{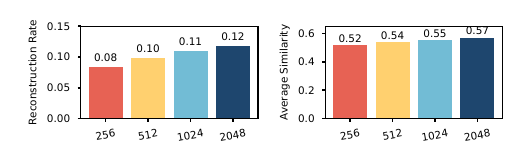}
\caption{Effect of fingerprint length (256, 512, 1024, and 2048) on reconstruction performance on molecules sampled from ChEMBL. The left panel shows the Reconstruction Rate, while the right panel displays the Average Similarity. Both metrics indicate that longer fingerprints enhance model performance. All models were trained for an identical wall time of 24 hours and used the same hyperparameters, except for the varying fingerprint lengths and the corresponding network widths.}
\label{fig:fp_length}
\end{figure}

The introduction of the diffusion module offers greater flexibility in handling longer fingerprint lengths. We trained models with varied fingerprint lengths (256, 512, 1024, and 2048 bits), all under identical conditions: a fixed training duration of 24 hours and consistent hyperparameters, except for adjustments in fingerprint length and the corresponding network widths. The models were evaluated on their reconstruction performance using ChEMBL samples, as shown in Figure \ref{fig:fp_length}. The results demonstrate that increasing the fingerprint length significantly enhances model performance, with a shift from 256 bits to 2048 bits resulting in an increase in the reconstruction rate from 8\% to 12\%, representing a 50\% improvement. These findings indicate that the diffusion module not only strengthens the modeling of molecular fingerprints but also scales effectively to longer fingerprints, enabling more precise nearest neighbor searches with higher resolution.

\clearpage
\subsection{Synthesizable analog generation}
This section provides additional results for synthesizable analog generation, focusing on two sets of experiments: (1) finding synthesizable analogs for \textit{de novo} design efforts using the GuacaMol benchmark \cite{brown2019guacamol} (where scores range from 0 to 1, with higher scores indicating better performance), and (2) generating molecules with Pocket2Mol \cite{luo20213d}, a structure-based design algorithm.

\begin{figure}[h!]
\centering
\includegraphics[width=0.75\textwidth]{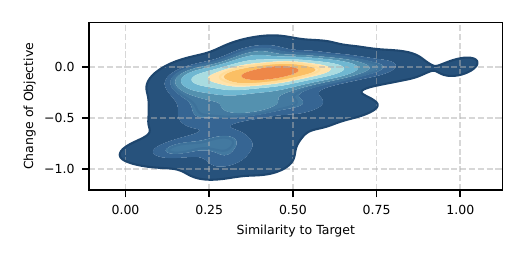}
\caption{Relationship between similarity to the target molecule and the change in objective value. The contour plot shows the density of data points, indicating that higher similarity to the target molecule is generally associated with neutral changes in the objective value. This suggests that closer matches to the target tend to maintain the main objective score. The distribution highlights that a significant fraction of outputs exhibit neutral changes in objective scores.}
\label{fig:goal_hard_cwo_unsynth_decode_distribution}
\end{figure}

Figure \ref{fig:goal_hard_cwo_unsynth_decode_distribution} presents the density plot of all molecules generated for unsynthesizable designs within the GuacaMol benchmark, plotted along the similarity to the target axis and the change in the main objective score axis. The results indicate that a substantial fraction of generated molecules exhibit relatively neutral changes in the main objective score, suggesting that the primary design objectives are largely preserved during analog generation. This is particularly evident for molecules with higher similarity to the target, which are clustered around zero change in the objective value, demonstrating that structurally similar analogs tend to retain their original objective scores.

However, the distribution also reveals a region characterized by low similarity and significant negative changes in objective scores, indicating instances where the model fails to generate meaningful analogs. These cases often result in outputs resembling random generation, highlighting the model’s limitations in some regions of the chemical space. This failure is partly due to the limited coverage of the explored chemical space compared to the theoretical coverage and partly due to the unsynthesizable input molecule being too distant from the synthesizable chemical space, making the “projection” unrealistic or irrelevant.

\begin{figure}[h!]
\centering
\includegraphics[width=0.75\textwidth]{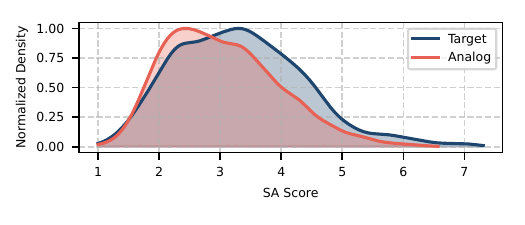}
\caption{Distribution of SA Scores for molecules designed using Pocket2Mol (denoted as Target) and their corresponding analogs (denoted as Analog). The shift of the Analog distribution towards lower SA Scores indicates that the generated analog molecules are generally easier to synthesize compared to the original designs from Pocket2Mol.}
\label{fig:sbdd_decode_sa_dist}
\end{figure}

Figure \ref{fig:sbdd_decode_sa_dist} illustrates the distribution of Synthetic Accessibility (SA) Scores for molecules designed using Pocket2Mol (labeled as Target) and their corresponding generated analogs (labeled as Analog). The plot reveals a noticeable shift in the Analog distribution towards lower SA Scores, suggesting that the analog molecules are generally easier to synthesize compared to the original Pocket2Mol designs.

\begin{figure}[ht!]
\centering
\includegraphics[width=0.85\textwidth]{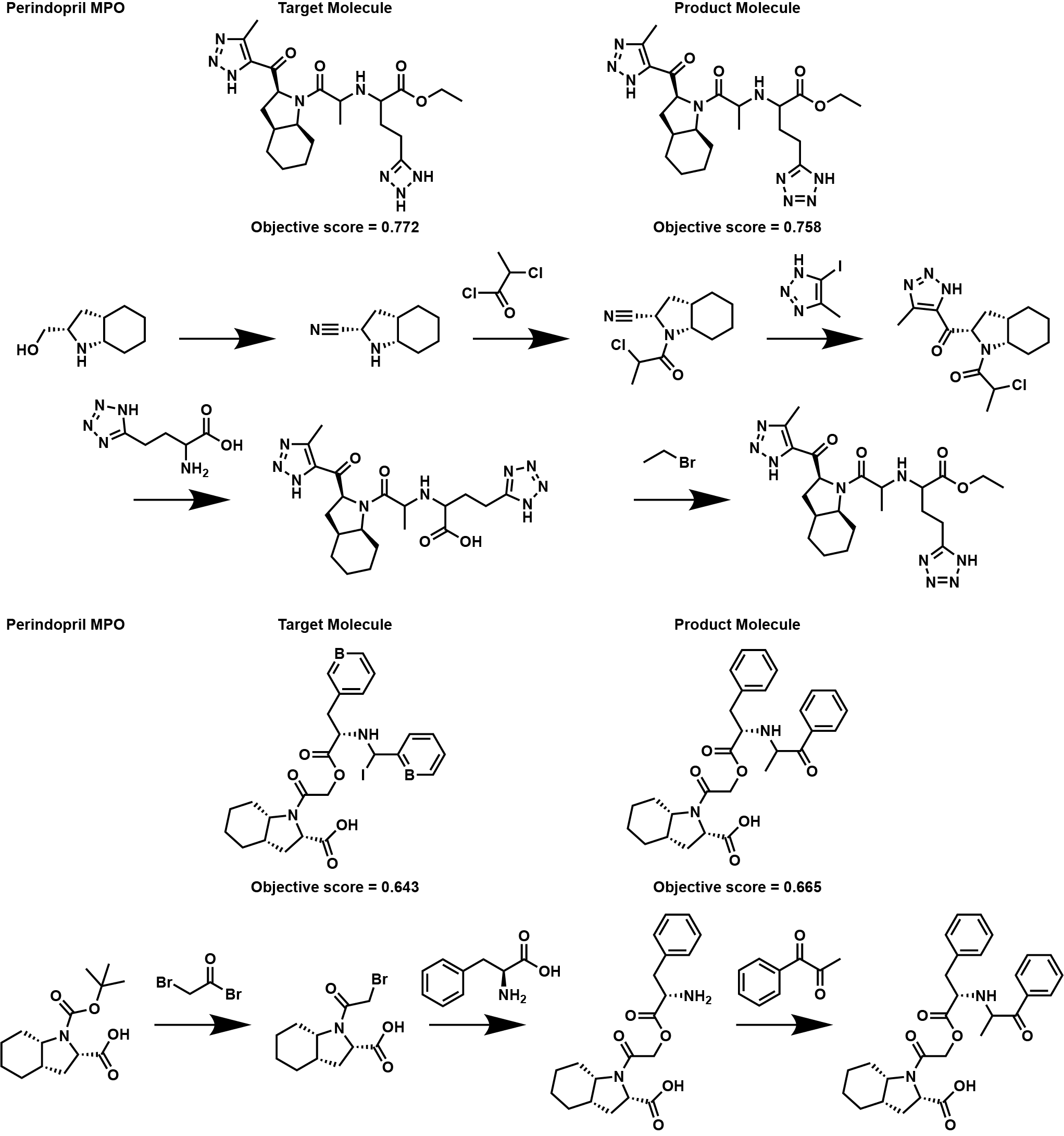}
\caption{Synthetic pathways for Perindopril MPO optimization examples as shown in the main text.}
\label{fig:unsyn_additional1}
\end{figure}

\begin{figure}[ht!]
\centering
\includegraphics[width=0.85\textwidth]{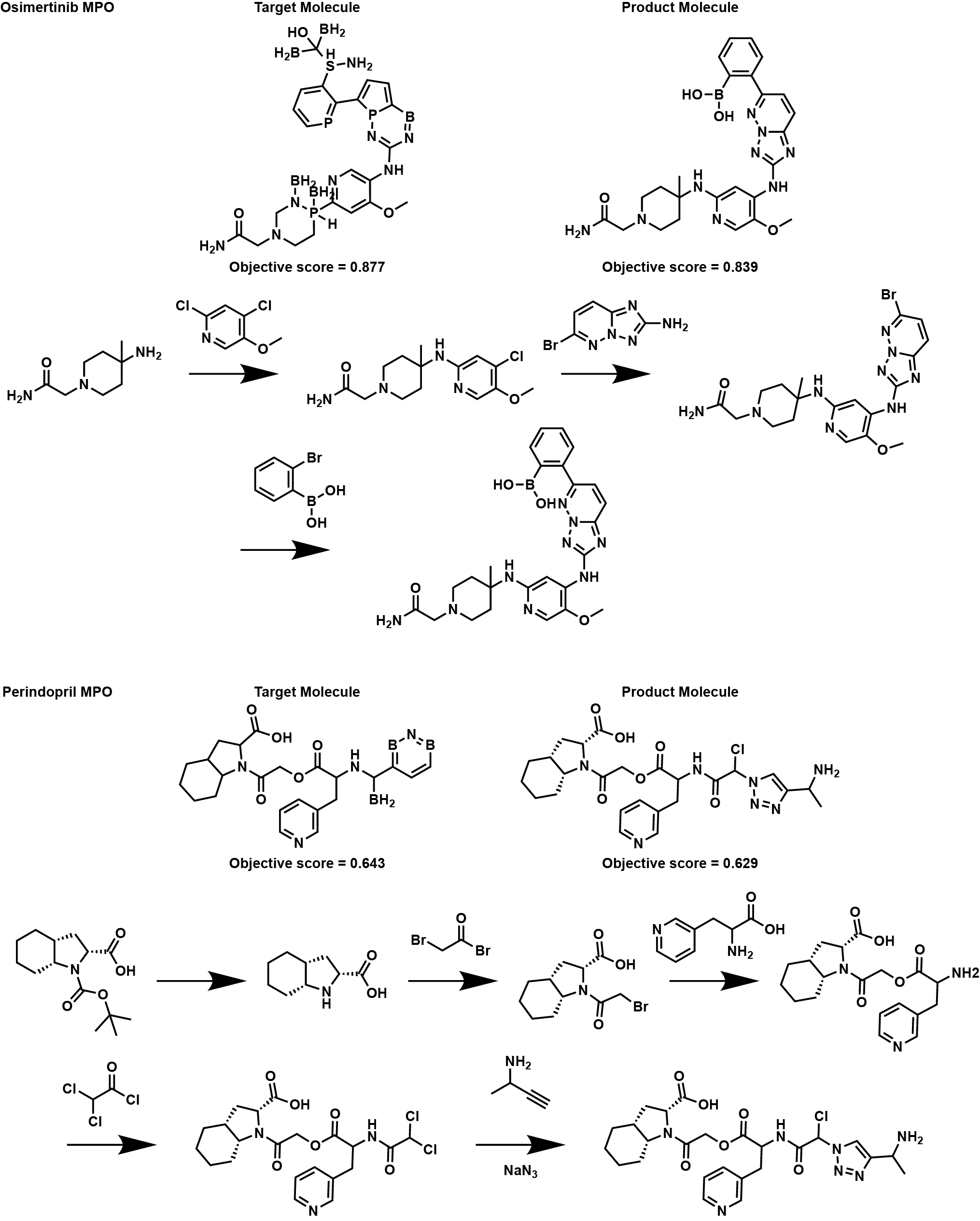}
\caption{Additional examples of synthesizable analog generation for previously unsynthesizable designs. Each panel shows the transformation from the target molecule (left) to the product molecule (right), with objective scores indicating the performance of each molecule in the optimization task. The synthetic routes for the synthesizable analogs are shown below each panel, demonstrating the practical feasibility of the proposed transformations.}
\label{fig:unsyn_additional2}
\end{figure}

\clearpage
\subsection{Hit expansion}
This section provides additional results for hit expansion experiments, including (1) expanding hits from screening ZINC250k with activity predictors and (2) expanding from experimentally verified binders of disease targets. 

\begin{figure}[h!]
    \centering
    \begin{subfigure}[b]{0.48\textwidth}
        \centering
        \includegraphics[width=\textwidth]{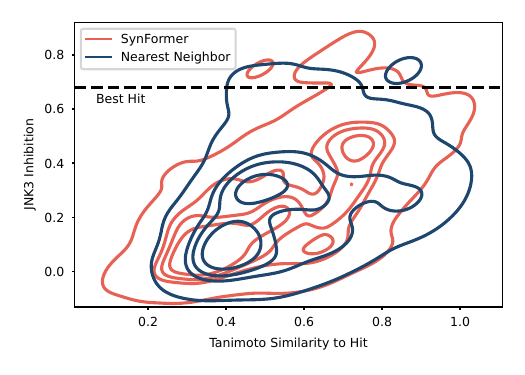}
        \label{fig:hit_expansion_jnk}
    \end{subfigure}
    \hfill
    \begin{subfigure}[b]{0.48\textwidth}
        \centering
        \includegraphics[width=\textwidth]{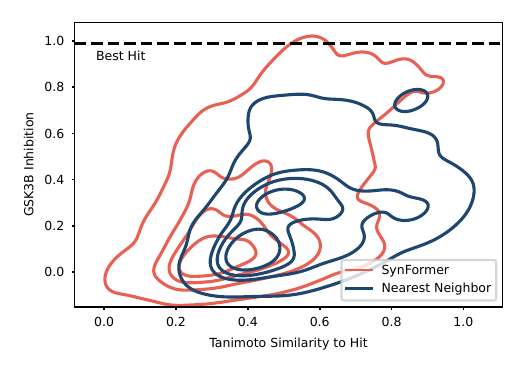}
        \label{fig:hit_expansion_gsk}
    \end{subfigure}
    \caption{Comparison of hit expansion performance between SynFormer (red contours) and Nearest Neighbor (blue contours) approaches for JNK3 (left) and GSK3$\beta$ (right). The contour plots show inhibition scores versus Tanimoto similarity to the input hit compound. The dashed line represents the inhibition level of the best hit compound.}
    \label{fig:combined_expansion_comparison}
\end{figure}

\begin{figure}[h!]
    \centering
    \begin{subfigure}[b]{\textwidth}
        \centering
        \includegraphics[width=\textwidth]{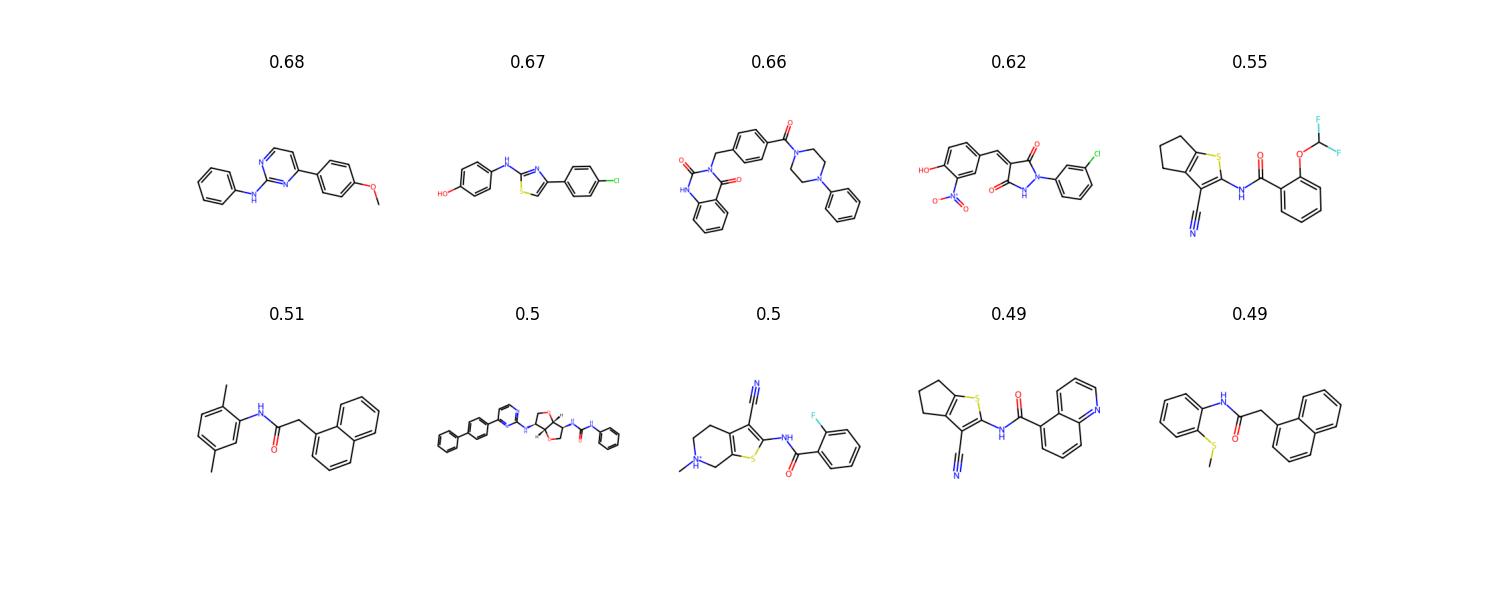}
        \caption{Hits for JNK3 identified from screening the ZINC250k database, with the predicted probability of being active labeled above each hit.}
        \label{fig:hits_jnk}
    \end{subfigure}
    \hfill
    \begin{subfigure}[b]{\textwidth}
        \centering
        \includegraphics[width=\textwidth]{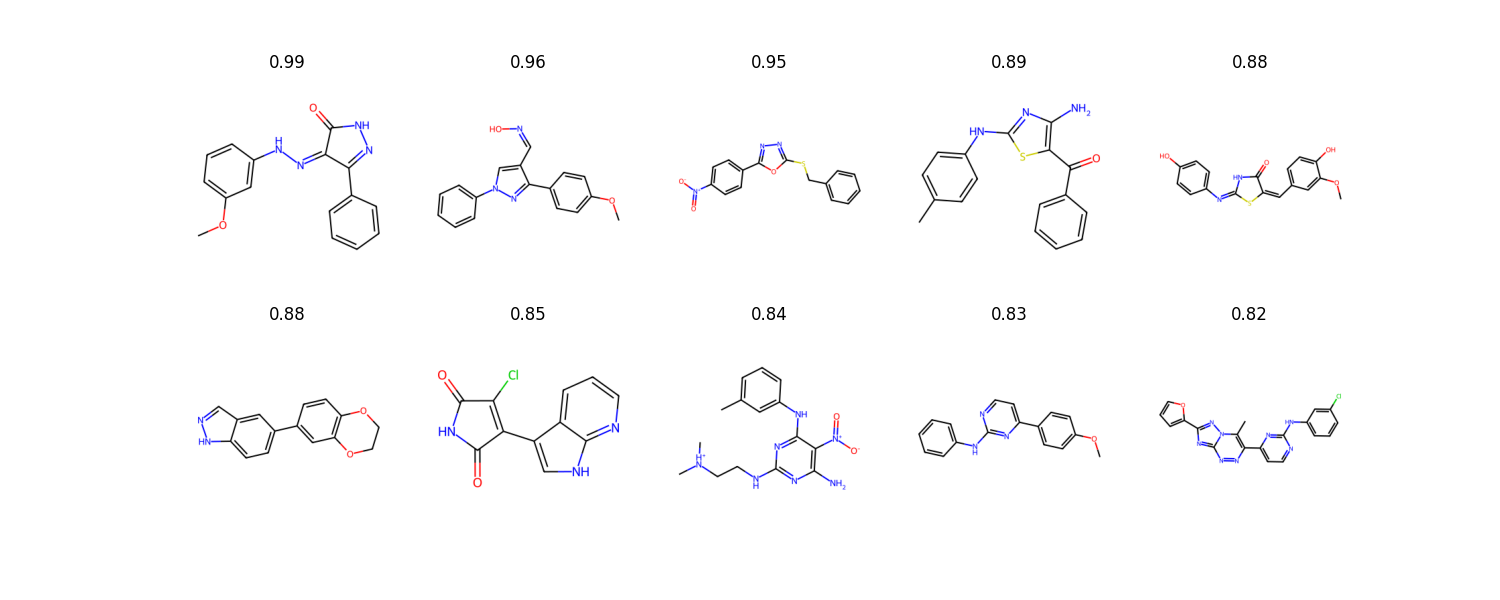}
        \caption{Hits for GSK3$\beta$ identified from screening the ZINC250k database, with the predicted probability of being active labeled above each hit.}
        \label{fig:hits_gsk}
    \end{subfigure}
    \caption{Hits identified from screening the ZINC250k database, which are subsequently used as inputs to the SynFormer for hit expansion.}
    \label{fig:hits}
\end{figure}

In addition to JNK3, which is discussed in the main text, we also evaluated SynFormer on GSK3$\beta$. Figure \ref{fig:combined_expansion_comparison} displays contour plots comparing the density of molecules generated by SynFormer with those identified by the nearest neighbor approach for both JNK3 and GSK3$\beta$. These plots illustrate the performance differences between the two methods in lead expansion. Additionally, Figure \ref{fig:hits} presents the top-10 molecules identified from screening the ZINC250k database using each activity predictor.

\begin{figure}[h!]
\centering
\includegraphics[width=0.5\textwidth]{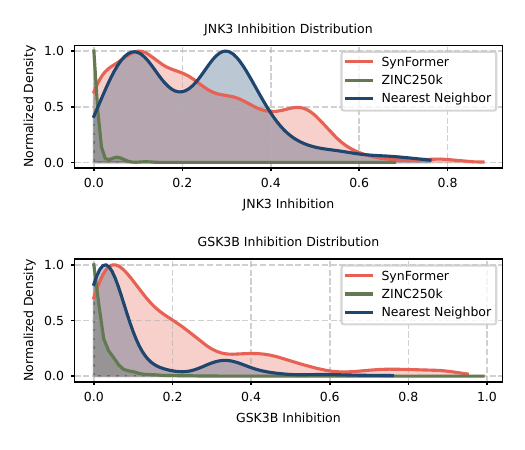}
\caption{Normalized density plots of inhibition scores for JNK3 (top) and GSK3$\beta$ (bottom) showing the performance of SynFormer (red), nearest neighbor search (blue), and the initial ZINC250k screening (green). SynFormer consistently generates molecules with higher inhibition scores compared to the nearest neighbor and ZINC250k hits, demonstrating its superior capability in identifying potent analogs for both targets.}
\label{fig:hit_expansion_score_dist}
\end{figure}

The density plots comparing analog discovery by nearest neighbor and SynFormer highlight that SynFormer consistently outperforms the nearest neighbor approach in identifying higher-scoring molecules in both tasks. As discussed in the main text, both SynFormer and nearest neighbor search identified molecules with predicted JNK3 inhibition scores higher than the best hits from initial screening, and SynFormer successfully found molecules with higher score than the highest one from the nearest neighbor search. For GSK3$\beta$, the nearest neighbor approach failed to find molecules approaching the highest inhibition values of the original hits, whereas SynFormer succeeded. This is also demonstrated in Figure \ref{fig:hit_expansion_score_dist}. Although nearest neighbor search showed enrichment in regions of higher similarity to the input references, molecules with scores exceeding those of the input references were often found at moderate similarity levels, ranging from 0.5 to 0.8. This observation suggests that SynFormer's ability to explore beyond close structural analogs allows it to identify more potent molecules while maintaining a reasonable balance between structural similarity and enhanced activity.

\begin{figure}[h!]
\centering
\includegraphics[width=0.85\textwidth]{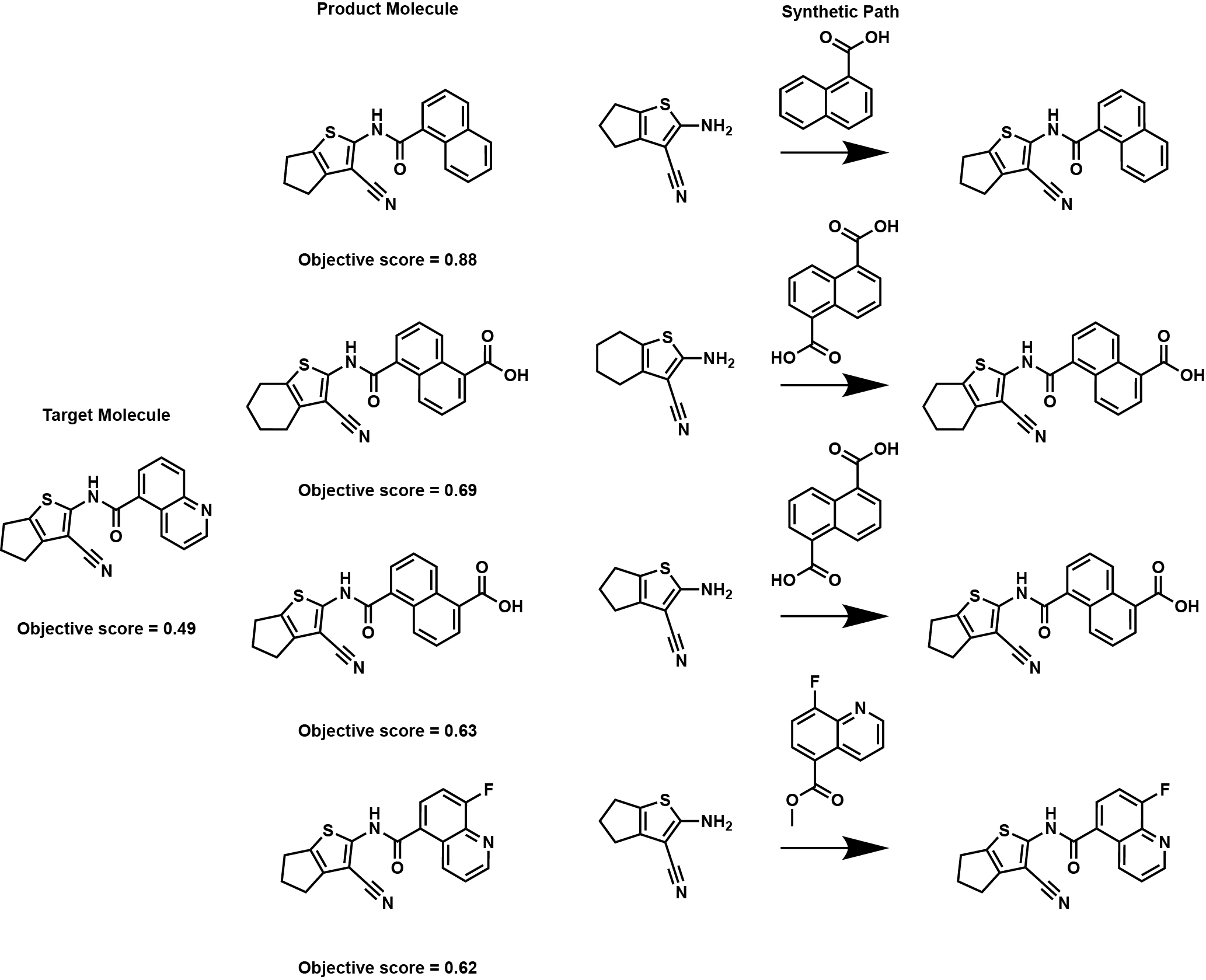}
\caption{Additional examples of molecules generated from the hit against JNK3, along with their predicted inhibition scores between 0 and 1 and corresponding synthetic pathways.}
\label{fig:jnk_analogs_si}
\end{figure}

\begin{figure}[h!]
\centering
\includegraphics[width=0.85\textwidth]{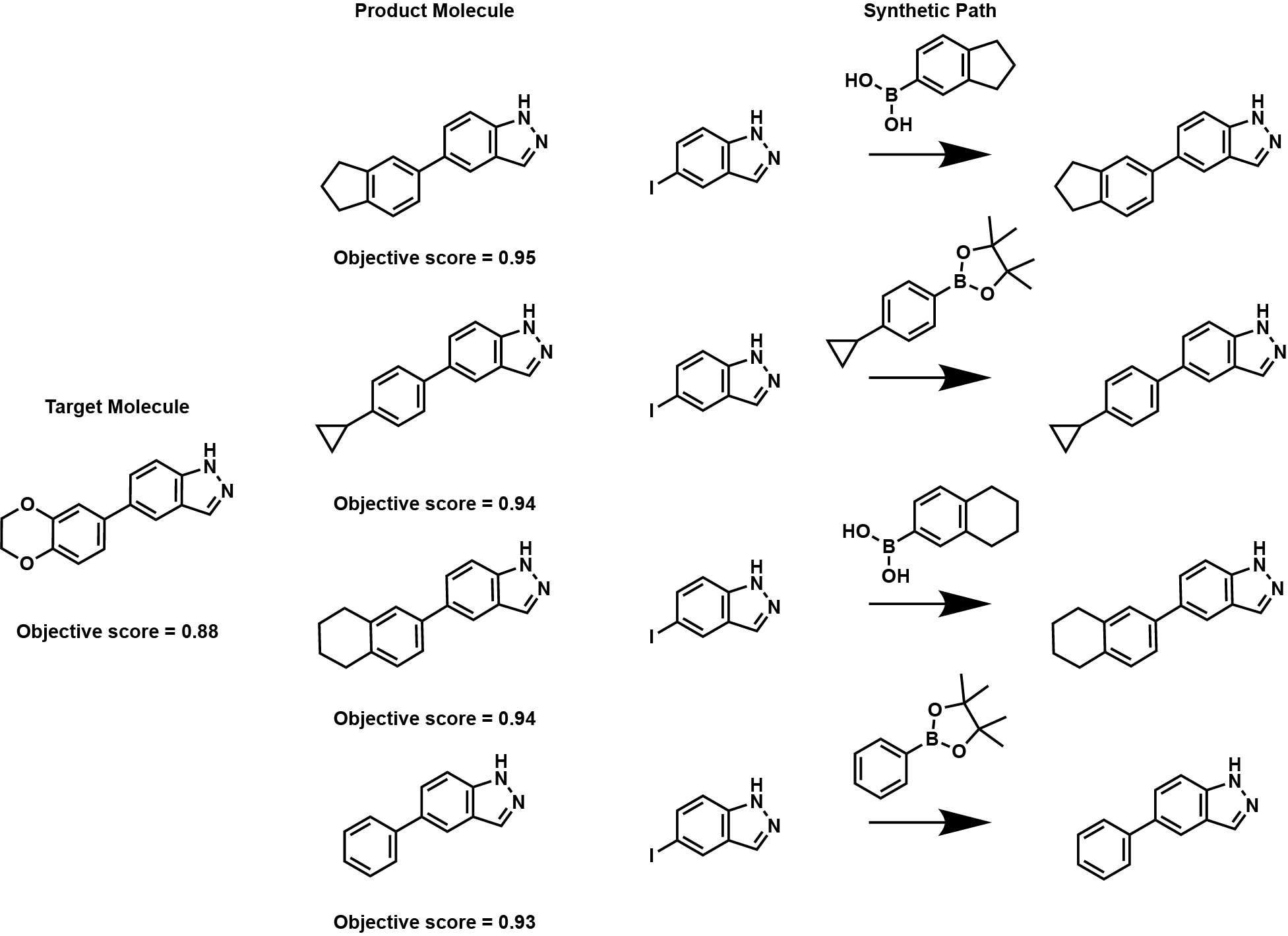}
\caption{Additional examples of molecules generated from the hit against GSK3$\beta$, along with their predicted inhibition scores between 0 and 1 and corresponding synthetic pathways.}
\label{fig:gsk_analogs_si}
\end{figure}

\begin{figure}[h!]
\centering
\includegraphics[width=\textwidth]{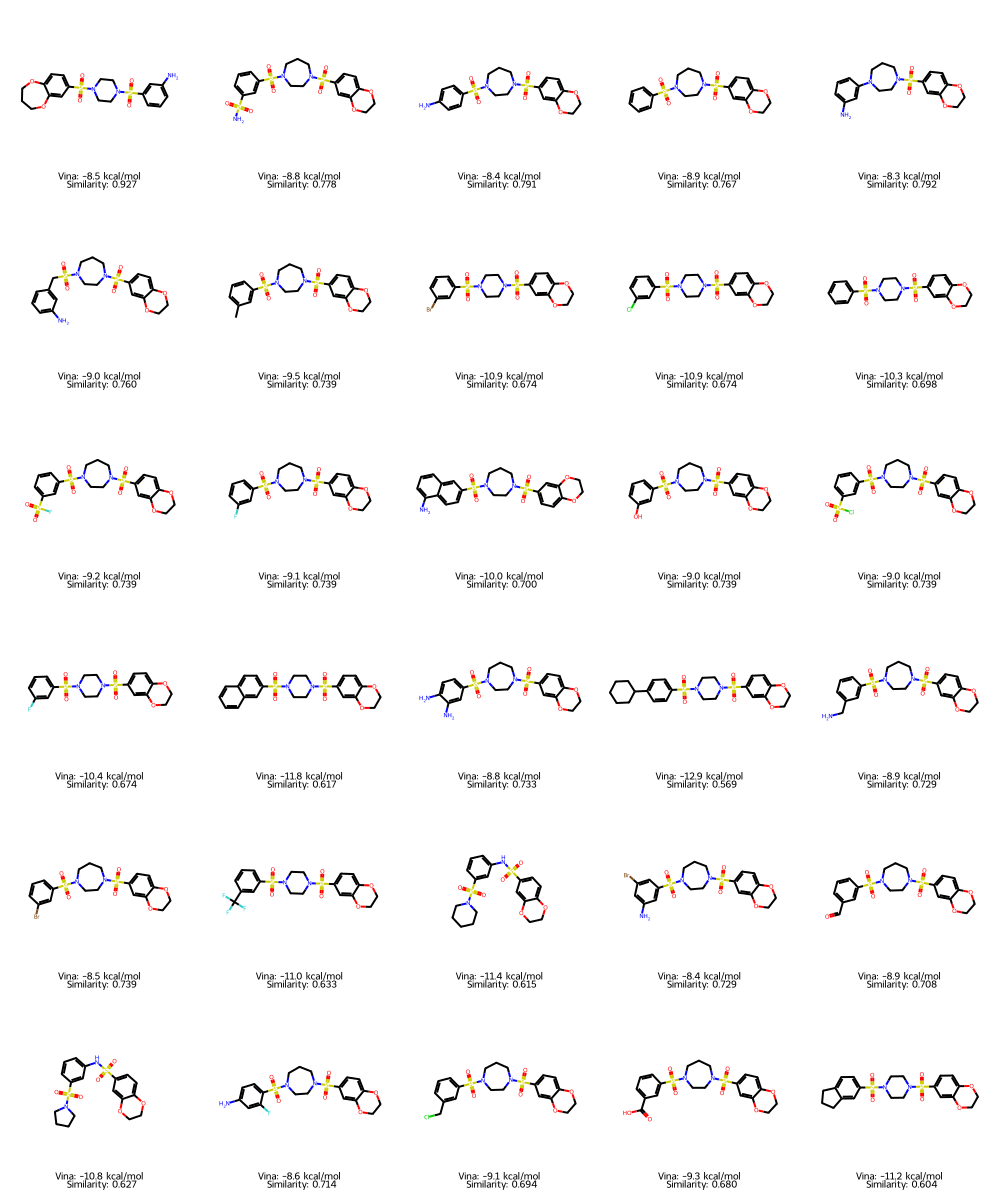}
\caption{Additional examples of molecules generated from the hit against PKM2.}
\label{fig:hit_exp_3me3}
\end{figure}

\begin{figure}[h!]
\centering
\includegraphics[width=\textwidth]{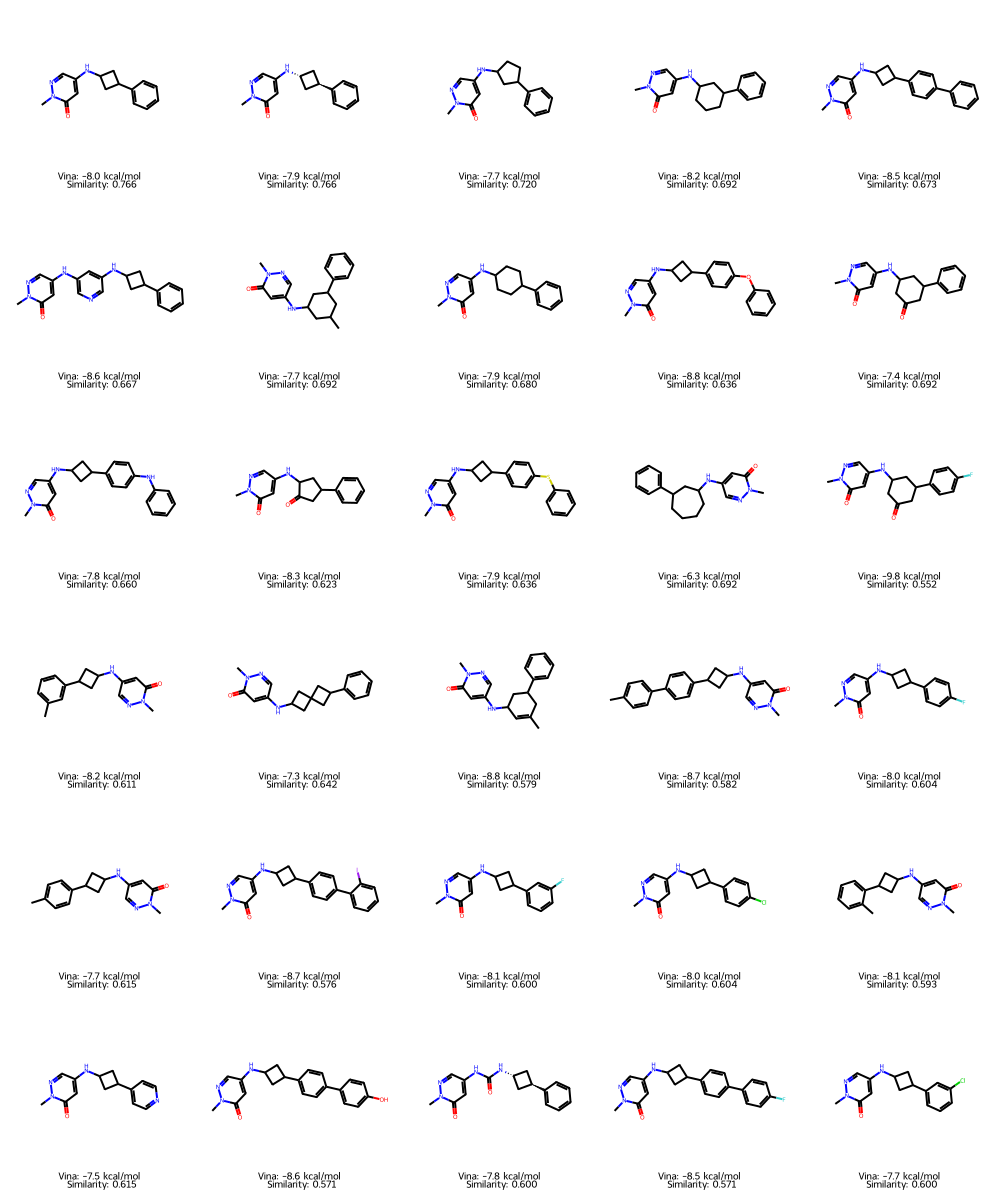}
\caption{Additional examples of molecules generated from the hit against KAT2A.}
\label{fig:hit_exp_5mlj}
\end{figure}

\clearpage
\subsection{Global chemical space exploration by \textit{de novo} molecular optimization}

\subsubsection{Molecular optimization using reinforcement learning}
This section describes additional results for using reinforcement learning to bias the generation of SynFormer-D toward high-scored molecules.

\begin{figure}[H]
    \centering
    \begin{subfigure}[b]{0.75\textwidth}
        \centering
        \includegraphics[width=\textwidth]{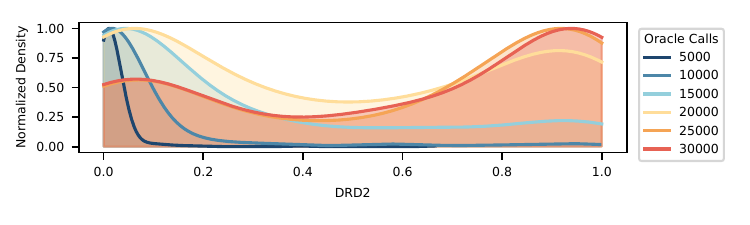}
        \caption{Score distribution of 200 molecules generated by SynFormer at different optimization steps, demonstrating that reinforcement learning effectively biases the generation distribution towards higher-scoring molecules.}
        \label{fig:rl_moving}
    \end{subfigure}
    \hfill
    \begin{subfigure}[b]{0.75\textwidth}
        \centering
        \includegraphics[width=\textwidth]{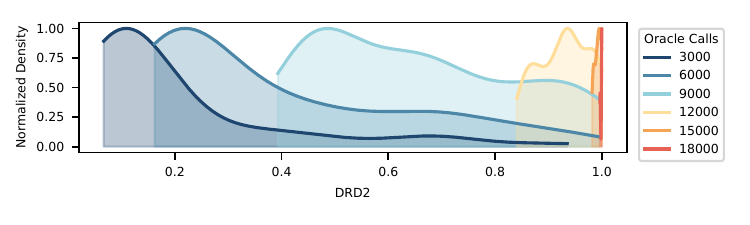}
        \caption{Score distribution of the accumulated top-100 molecules at various optimization steps, highlighting the progression and refinement of high-scoring molecules throughout the optimization process.}
        \label{fig:rl_accumul}
    \end{subfigure}
    \caption{Additional results for SF-RL, the fine-tuned version of SynFormer-D using an approach adapted from REINVENT.}
    \label{fig:rl}
\end{figure}

Figure \ref{fig:rl_moving} presents the score distribution of molecules generated by the SynFormer model after exposure to 5,000, 10,000, 15,000, 20,000, 25,000, and 30,000 labeled molecules and their corresponding scores. As the model processes more score-labeled data, the generation progressively shifts toward higher-scoring regions, demonstrating the effectiveness of the reinforcement learning (RL) algorithm in guiding the model toward optimized outputs. Figure \ref{fig:rl_accumul} shows the accumulation of the top-100 molecules throughout the optimization process. The results indicate that the model's performance saturates at the maximum score of around 1.0 after approximately 18,000 oracle evaluations, confirming the RL approach's ability to consistently identify top-performing molecules.

\subsubsection{Molecular optimization using genetic algorithm}
This section describes additional results for using a genetic algorithm with SynFormer-ED as a mutation.

\begin{figure}[H]
\centering
\includegraphics[width=0.65\textwidth]{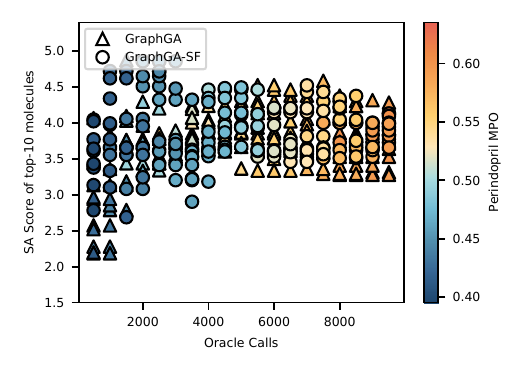}
\caption{Distribution of SA Scores \cite{ertl2009estimation} for the top 25 molecules at various optimization steps, with colors representing objective scores.}
\label{fig:comb_ga_per_sa}
\end{figure}

\begin{figure}[H]
\centering
\includegraphics[width=0.65\textwidth]{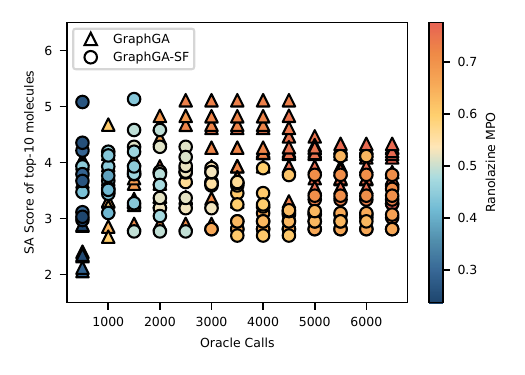}
\caption{Distribution of SA Scores \cite{ertl2009estimation} for the top 25 molecules at various optimization steps, with colors representing objective scores.}
\label{fig:comb_ga_ran_sa}
\end{figure}

\begin{figure}[H]
\centering
\includegraphics[width=0.65\textwidth]{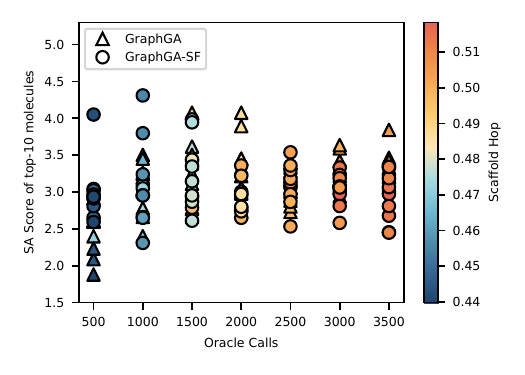}
\caption{Distribution of SA Scores \cite{ertl2009estimation} for the top 25 molecules at various optimization steps, with colors representing objective scores.}
\label{fig:comb_ga_sca_sa}
\end{figure}

Figure \ref{fig:comb_ga_per_sa}, \ref{fig:comb_ga_ran_sa}, and \ref{fig:comb_ga_sca_sa} display the distribution of SA Scores \cite{ertl2009estimation} for the top 25 molecules at various optimization steps during the optimization of Perindopril MPO, Ranolazine MPO, and Scaffold Hop, respectively. In these tasks, since the main objective functions do not penalize synthetic accessibility, the differences between the standard GraphGA and GraphGA-SF are less pronounced compared to those observed in the Sitagliptin MPO task discussed in the main text.

\end{document}